\documentclass[3p, review]{elsarticle}

\usepackage{easyReview}
\usepackage{amsmath,amssymb}
\usepackage{color}
\usepackage{multirow}
\usepackage{paralist}
\usepackage{filinStyle}
\usepackage{diagbox}
\usepackage{import}
\usepackage{threeparttable}
\usepackage{url}

\def\x{{\bf x}}

\graphicspath{{figures/}}

\begin{document}

\title{Evaluating saliency scores in point clouds of natural environments by learning surface anomalies}


	\author[1,2]{Reuma Arav\corref{cor1}}\ead{reuma.arav@boku.ac.at}
	\author[3]{Dennis Wittich}\ead{wittich@ipi.uni-hannover.de}
	\author[3]{Franz Rottensteiner}\ead{rottensteiner@ipi.uni-hannover.de}

	
	\affiliation[1]{organization={Dept. of Geodesy and Geoinformation, TU Wien},
	addressline={Wiedener Hauptstrasse 8},
	city={Vienna},
	postcode={1040},
	country={Austria}
	}
	\affiliation[2]{organization={Institute of Geomatics, University of Natural Resources and Life Sciences, Vienna (BOKU)},
	addressline={Peter-Jordan-Strasse 82},
	city={Vienna},
	postcode={1190},
	country={Austria}
}
	\affiliation[3]{
	organization={Institute of Photogrammetry and GeoInformation, Leibniz University Hannover},
	city={Hannover},
	country={Germany}}
	
	\cortext[cor1]{Corresponding author}


\begin{abstract}
In recent years, three-dimensional point clouds are used increasingly to document natural environments. 
Each dataset contains a diverse set of objects, at varying shapes and sizes, distributed throughout the data and intricately intertwined with the topography. 
Therefore, regions of interest are difficult to find and consequent analyses become a challenge. Inspired from visual perception principles, we propose to differentiate  objects of interest from the cluttered environment by evaluating how much they stand out from their surroundings, i.e., their geometric salience. 
Previous saliency detection approaches suggested mostly handcrafted attributes for the task. However, such methods fail when the data are too noisy or have high levels of texture.  
Here we propose a learning-based mechanism that accommodates noise and textured surfaces. 
We assume that within the natural environment any change from the prevalent surface would suggest a salient object. 
Thus, we first learn the underlying surface and then search for anomalies within it. 
Initially, a deep neural network is trained to reconstruct the surface. Regions where the reconstructed part deviates significantly from the original point cloud yield a substantial reconstruction error, signifying an anomaly, i.e., saliency.
We demonstrate the effectiveness of the proposed approach by searching for salient features in various natural scenarios, which were acquired by different acquisition platforms. 
We show the strong correlation between the reconstruction error and salient objects.

\end{abstract}

\maketitle   

\section{INTRODUCTION}\label{introduction}
 
 \newcommand{\hlcite}[0]{(\hl{cite})}
 
Three-dimensional point clouds have become an essential tool for geoscientific studies. 
Everything within the natural environment is being documented and monitored: from millimetre-wide cracks, to centimetre-long blocks and metre-wide rivers  \citep{Telling.etal2017,Tarolli.Mudd2020,Kyriou.etal2021}.   
The acquired point clouds provide a high resolution description of the landscape, enabling analyses that would otherwise be impossible.
These datasets are characterized by a massive amount of unorganized points, which span over wide areas at different point spacing. 
The collected data comprise a diverse array of objects of interest with varying shapes and sizes, distributed throughout the dataset and embedded within the topography.
Due to  acquisition conditions, the data hold a significant amount of noise and uninteresting regions make up a larger portion of the point cloud \citep{Arav.etal2022}. 

Studies have shown that focusing on important regions within the point cloud improves scene understanding \citep{Alexiou.etal2019, Liang.etal2023}. 
This can be accomplished through \textit{visual saliency}, which is defined as the subjective quality that makes certain objects or regions stand out in their environment, capturing the observer's attention \citep{Akman.Jonker2010}. 
In 3D, saliency is defined as objects (or regions) that stand out from their surroundings, also geometrically. 
Common saliency approaches in 3D point clouds focus on small object models (e.g., \citep{Guo.etal2018,Alexiou.etal2019,Ding.etal2019,Leal.etal2019}), where the point cloud is confined, resolution is approximately constant, and noise levels are often low. 
Studies that wish to extend the detection to larger scenes usually focus on urban environments. 
There, salient objects hold distinct features, so that first-order features, such as normal, height, or orientation are sufficient for saliency detection \cite{Hao.etal2019, Yun.Sim2016, Fan.etal2022}. 
However, these approaches fail in natural environments, where entities transform smoothly into  the background. 

In this paper we introduce a new approach to estimate saliency  in 3D point clouds of natural environments. 
To do so, we estimate  anomaly probability within a  surface. 
Based on the fact that landscapes are generally continuous and smooth, salient features will present an unexpected change in the surface. 
We propose to use a deep neural network to predict small parts of the landscape providing only a reduced amount of information. Then, we interpret the deviation between the actual and the predicted  surface as a measure of saliency for that area.
Specifically, we train a network by inputting the outer cells (a shell) of a voxelized region (voxel grid) and generating a predicted voxel grid as output.
It is assumed that the shell contains all the required information to predict the surface described by the voxel grid, as long as the inner part is regular. 
However, whenever the inner part is irregular, the reconstruction error will be large, and thus will signify high saliency.
We demonstrate the proposed approach in three real-world settings, that substantially differ one from the other.
We show, both visually and quantitatively, the strong link between the reconstruction error and salient objects. 
Doing so, we propose a new approach for evaluating saliency in 3D point clouds, which, unlike current deep learning approaches, does not consider saliency detection as a classification problem. Therefore, it does not require pre-trained classifiers. 
By predicting the surface from the obtained point cloud, the proposed approach can detect saliency in open terrain datasets and is not limited to small objects. 
Furthermore, is can handle substantial data volumes, high noise levels, and irregular point distribution, all of which are inherent characteristic to 3D point clouds acquired by laser scanning platforms. 
To promote further study of saliency estimation algorithms, we release our source code \cite{code2023}. 
%
%

\section{Related work}\label{related_work}

Saliency detection in 3D point clouds has been gaining popularity for several years as a preliminary process for various complex processing tasks. For example, \citet{Qin.etal2023} use saliency to register multiple point clouds  of an indoor scene,  \citet{Laazoufi.Hassouni2022} use salient points to evaluate point cloud quality,
\citet{Liang.etal2023} enhance point cloud models by reducing excessive non-salient points that obscure the overall shape of the model, and
\citet{Hong.etal2023} employ salient regions in data augmentation learning models for segmentation.
 
Saliency detection approaches in full 3D data are quite rare. Nonetheless, there are many works that deal with saliency detection in RGB-D (colour and depth) images. There, saliency is found mostly based on RGB information, while the depth map is used to improve results. 
In recent years, most methods adopt deep learning models for the task. 
\citet{Chen.etal2021a} and \citet{Zhou.etal2021} differentiate between early, late, and middle aggregation approaches. 
In  early aggregation models, both RGB and depth images are fused in the input level, and then a CNN-based network is used to extract the features for saliency detection \citep[e.g.,][]{Zhang.etal2020a, Zhang.etal2021}. 
In late aggregation approaches, saliency cues are learned separately from the depth and colour channels before being fused to obtain the saliency map \citep[e.g.,][]{Li.etal2022, Sun.etal2022, Chen.etal2021}. 
For example, \citet{Chen.etal2021} learn the relevant cues for saliency detection from each channel, and then select cues that exist only in one channel. The saliency inference is carried out by fusing low- and high-level cues from both channels. 
Middle aggregation models try to combine both early and late aggregation approaches, so that learning is carried out in two phases. In the first phase, saliency features are obtained for each modality. In the second phase, they are fused to generate the final saliency map \citep{Chen.etal2021a, Zhou.etal2021, Han.etal2018, Zhang.etal2021a}. 
For instance, \citet{Zhou.etal2021} first feed the depth and colour images into two learning networks to obtain corresponding multi-level feature representations. These representations are fused using an integration module, where a shared learning network enhances the features for saliency detection. 

However, the works above capitalize on existing saliency approaches in colour images, while assuming a corresponding depth map. 
Yet, colour information in  3D point clouds  is not always available,  necessitating a greater emphasis on geometric features. 
Moreover, the data is unorganised, with varying point spacing, and in three dimensions, making such raster-based approaches inapplicable. 
Works that define saliency particularly in point clouds are rare. Still, we divide them here to handcrafted and deep learning based approaches.

\subsection{Handcrafted saliency approaches} 
\citet{Shtrom.etal2013} were first authors to introduce saliency in point clouds which completely relies on geometric characteristics. The authors computed a fast point feature histogram descriptor (FPFH, \cite{Rusu.etal2009})  and then evaluated its distinction from the local neighbourhood. A global rarity was then estimated by measuring  the dissimilarity between every two points in the cloud. This approach was applied successfully to both small object models and urban environments in other works \citep[e.g.,][]{Kobyshev.etal2016, Hao.etal2019}. \citet{Tasse.etal2015, Yun.Sim2016} and also \citet{Ding.etal2019} improved its  computational efficiency by using cluster-wise comparison rather than a point-wise one.    
Other approaches for saliency detection proposed to use different metrics of local distinctness. Nonetheless, these were also based on normal computation and the distinction of the point's normal from its immediate surrounding. 
\citet{Wang.etal2015} measured the difference of a point's normal from the dominant normal in the scene. Applied to roads scanned by mobile scanners, this approach is  aimed specifically to highlight off-road objects.   
\citet{Guo.etal2018} defined a point descriptor based on principal component analysis (PCA). 
The descriptor was composed of sigma-sets extracted from the covariance matrix of each point’s normal and curvature. 
\citet{Arvanitis.etal2022} defined salient points as those belong to non-flat surfaces. 
The flatness is determined by the covariance matrix eigenvalues of a local neighbourhood. Non-flat areas produce low eigenvalues that correspond to high saliency values. 
In such normal-based approaches, the assumption is that a salient feature is defined by an abrupt change in orientation. However, in natural environments this might not be the case. There, entities such as gullies, landslides, rockfalls, sinkholes, or cracks, are parts of the underlying surfaces. 
Such objects have intermediate borders, which gradually and continuously change from background to entity \citep{Molenaar.Cheng2000, Liu.etal2019}. Therefore, though they differ from their surroundings, their borders are mostly vague and are hard to define  \cite{Molenaar.Cheng2000}.  
To overcome this problem, \citet{Arav.Filin2020} proposed a method that is attuned to detect vague objects as salient features. Instead of looking for an immediate change in the local surrounding, the authors suggest to look at a farther neighbourhood. Furthermore, to allow for more subtle objects, the authors do not only take the normal change into account, but also the change rate, i.e., the curvature. 
The advantage of this approach was shown in later works \cite{Arav.Filin2022, Arav.etal2022a}  detecting salient objects in different types of natural scenes, including a complete 3D scenario (i.e., a cave). 
Nonetheless, this approach would fail in cases of rough surfaces (e.g., riverbeds, alluvial fans). There, the difference between a point and its wider surrounding is high, leading to an increased sensitivity in detection. Moreover, outliers (i.e., measurement noise) will also be highlighted, as their normals and curvatures completely differ from their surroundings. 

The review has shown that handcrafted approaches for saliency estimation evaluate how much a point differs from its surrounding \citep[aka. centre-surround principle][]{Itti.etal1998}, mostly focusing on the difference in normal direction. In such schemes, a larger context of  salient features is missing, leading to high sensitivity to local variations.

\subsection{Deep learning-based saliency approaches} 
To the best of our knowledge, only a few approaches were proposed for saliency evaluation in point clouds using deep learning.
These tend to use pre-trained models and are mostly in the context of shape recognition and classification. 
 \citet{Zheng.etal2019} assert that salient points explicitly explain which points are key for model recognition. The authors  assume that  points that lie on the object's borders contribute more to shape recognition than those that lie on its inner surface. 
Therefore, they suggest that elimination of unimportant points or their movement towards the object's inner surface are equivalent. Under this assumption, salient points are marked by the change in prediction loss using a pre-trained classifier for shape recognition. 
The change in prediction loss is approximated by the gradient of the loss when shifting points to the centre of the object's point cloud. These gradients were interpreted as  saliency scores. 
Another semi-supervised approach was proposed by \citet{Jiang.etal2023}. The authors use objects that were previously classified in order to learn the saliency. This is carried out in two main branches: a classification branch, which  uses category labels for feature extraction, and a saliency branch that uses a multi-scale point cluster matrix to provide coherent saliency regions.  
Both approaches target point clouds of objects whose category labels are known. 
Within the natural environment, where  objects may be restricted only to one region or may appear only a few times, training data for classification may be difficult to acquire.  Moreover, manual labelling which marks \textit{salient} and \textit{non salient} features in scenes as large as point clouds of natural environment are, is not only time-consuming, but also prone to perception bias and degrades the detection accuracy \citep{Hillier.etal2014, Scheiber.etal2015, Vinci.etal2016}. Therefore, the aforementioned methods cannot be applied to point clouds of natural environment. 
To overcome this problem, we propose a new approach to highlight salient regions that is independent of previous classification. The detection is driven by the notion that in natural environments salient object are in a way an anomaly in the general surface.

\section{Methodology}\label{methodology}
We seek to highlight salient features in point clouds, focusing on datasets that document natural environments (i.e., non-urban scenes). 
Following the notion that salient features are a sudden change in the surface, we assume that they  will present an irregularity at that location. 
Therefore, we consider the task of highlighting saliency as marking anomalies in the scene. 
To do so, we first train a deep neural network to reconstruct the surface from a reduced subset of the data. Then, we reconstruct the surface and evaluate the reconstruction error. This error is interpreted as the saliency score, since the reconstruction error will be larger in irregular regions.
In this way, we highlight salient features in 3D point clouds where external information is used only to find the best hyper-parameters of the method (i.e., hyper-parameters tuning). 

We begin with the details of our proposed method to mark salient regions \PSref{sec:method_general}. This section also includes   a formal definition of the problem and the notations used in this work. Then, we outline the network architecture \PSref{sec:architecture}, followed by details concerning the loss function and training procedures \PSref{sec:training}. Lastly, we describe how  saliency scores are estimated  \PSref{sec:inference}.

\subsection{Saliency estimation in 3D point clouds by anomaly detection}\label{sec:method_general}

Let $P$ be a point cloud, defined as a set of $N$ 3D points. These compile the main input to the method.
Additionally, as an input, we introduce two subsets of $P$: $H$, which is composed of points that are expected to have high saliency scores; and $L$, composed of points that are expected to have lower saliency scores. 
These subsets are required to tune the hyper-parameters of the method. Note that $H \cup L \neq P$.  They are only samples of each group $H$ and $L$. 
The output of the method is a saliency map, where each point $p_i \in P$ has a   saliency score $\xi_i$.

The saliency score is in fact an interpretation of the reconstruction error. It is  obtained  by a reconstruction network $R$ for a point's surrounding region. 
This network is pretrained  on random regions extracted from $P$ to enable a reconstruction of the surface recorded by the point cloud. This is based on the assumption that salient regions are rare, and therefore, the network will not learn them, but it will rather learn  regular surfaces. 
This approach has a major advantage, as we do not require any manually generated reference to train $R$. Instead, we use arbitrary sub-regions of $P$ for the task.

To evaluate the reconstruction error as a saliency score, one has to formulate the  reconstruction task in a way that  $R$ could reconstruct the surroundings of a point, as long as these surroundings are regular.
Yet, if the surroundings are irregular, the reconstruction should be incorrect, yielding a high reconstruction error, i.e., a high saliency score.
To do so, we use a voxel-based representation of the point cloud. Then, we formulate the reconstruction task to predict the inner part of a voxel grid based on its  outer voxels (the grid's shell).

Let $V_i$ be the representation of a region in $P$ in terms of a voxel grid of size $n\times n \times n$ that contains the surrounding region of $p_i$, such that $V_i$ is centred at $p_i$.
The side-length of each voxel cell in $V_i$ is parameterised by $w$, resulting in a volume of $w \times w \times w$ for each voxel cell and a total volume of $(w \cdot n)\times (w \cdot n)\times (w \cdot n)$ for $V_i$.
The value of a voxel $V_{i,(\hat{x},\hat{y},\hat{z})}$ at voxel coordinates ($\hat{x},\hat{y},\hat{z}$) in $V_i$ corresponds to the number of 3D points in that cell.

Next, we introduce $S_i$. This is a modified version of $V_i$, where the values in the inner cells are set to zero.
Consequently, $S_i$ contains only the information from the shell of $V_i$.
In particular, the value of the voxel cell $S_{i,(\hat{x},\hat{y},\hat{z})}$ is $V_{i,(\hat{x},\hat{y},\hat{z})}$ if $\min(\hat{x},\hat{y},\hat{z}) \leq m$ or $\max(\hat{x},\hat{y},\hat{z}) \geq n-m-1$, and zero otherwise.
Here, $m$ denotes the thickness of the shell, i.e., how many voxels compile the shell.
Using this notation, the reconstruction task is carried out by $R$. The network  predicts the values of a voxel grid $\hat{V}_i$ based on the shell $S_i$, such that $\hat{V}_i$ is similar to $V_i$.
To measure the similarity between $\hat{V}_i$ and $V_i$, we introduce a function $\mathcal{R}(\hat{V}_i, V_i)$ that measures the reconstruction error.
The selection of the reconstruction error function $\mathcal{R}$ is be discussed in \Sref{sec:training}.

\begin{figure}[h]
	\centering
	\includegraphics[width=0.7\linewidth]{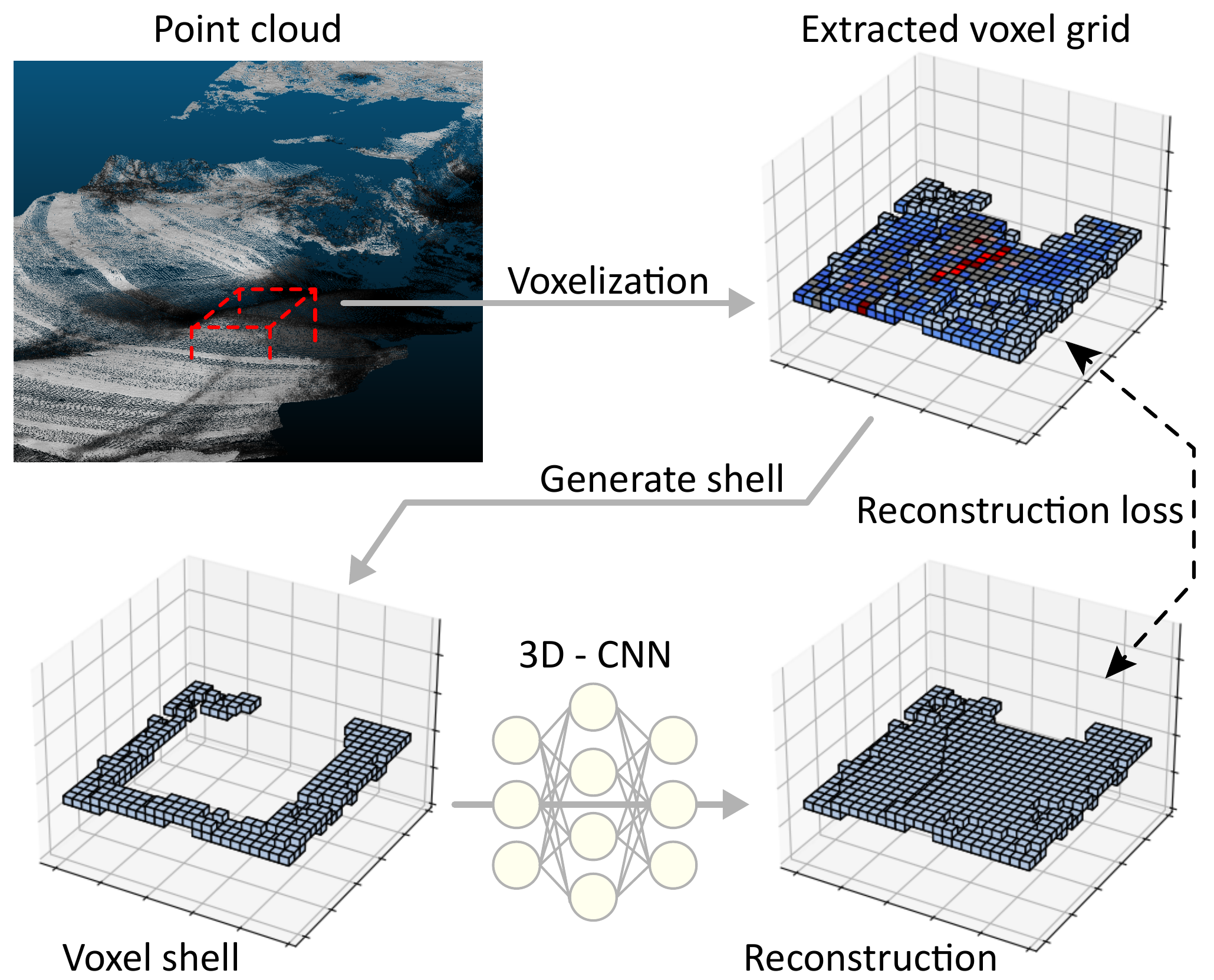}
	\caption{Overview of the training scheme. We generate random voxel grids from a set of given point-clouds. Next, the shells are generated by setting the inner voxels to zero. The task of the CNN is then to reconstruct the original input. Note that the extracted voxel grid is coloured by the number of points in each cell.  This information is used to calculate a weighted reconstruction loss.}
	\label{fig:method}
\end{figure}

The overall training scheme of our proposed method is shown in \Fref{fig:method}.
To train the reconstruction network $R$, we randomly select points from the cloud $P$, then voxelize their surrounding, resulting in voxel grids $V_i$. 
Based on these grids we generate corresponding shells $S_i$.
Then, the parameters of $R$ are obtained by minimizing the reconstruction error $\mathcal{R}(\hat{V}_i, V_i)$.
Eventually, the saliency score $\xi_i$ for a point $p_i$ is estimated by the reconstruction error $\mathcal{R}(\hat{V}_i,V_i)$ for each point in $P$ using the trained network. 
It should be mentioned that as the network is trained to reconstruct `regular' surfaces, and in each scene this `regularity' is defined differently,  training has to be conducted for each new scene. 


\subsection{Architecture}\label{sec:architecture}

\begin{figure}[h]
	\centering
	\includegraphics[width=0.9\linewidth]{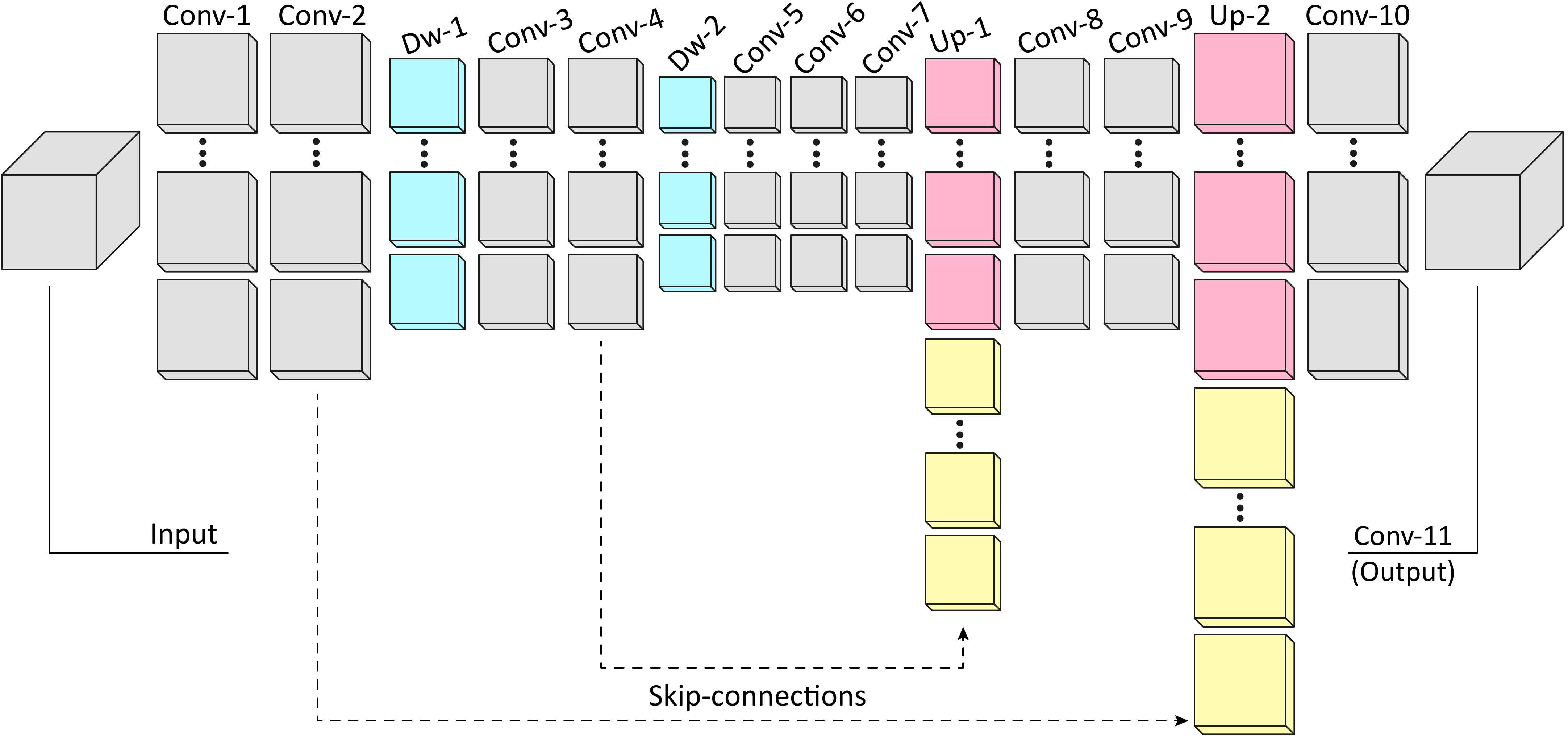}
	\caption{Illustration of the variational auto-encoder used to reconstruct the surface based on the information in the shell of the voxel grid representation. Blue: Output of downsampling. Red: Output of upsampling. Yellow: Concatenated feature tensors. }
	\label{fig:3d-unet}
\end{figure}

To perform the reconstruction task, we use a 3D convolutional neural network (CNN) as the reconstruction network $R$.
The architecture is shown in \Fref{fig:3d-unet}.
$R$ takes a shell $S_i$ as an input and outputs the values of a reconstructed voxel grid $\hat{V}_i$.
All tensors $S_i$, $V_i$ and $\hat{V}_i$ have the same shape, which is $n \times n \times n$, where $n$ is the side-length of the voxel grid.

The architecture of $R$ follows the encoder-decoder scheme with skip connections, similar to the U-Net architecture \cite{ronneberger2015}.
In this layout the spatial size of the feature maps is halved in each stage on the decoder and doubled in the decoder (cf. \Fref{fig:3d-unet}).
An illustration of the architecture is shown in \Fref{fig:3d-unet} with the corresponding layers described in \Tref{tab:arch}.

The capacity of the network is parametrized by the parameter $f$, which describes the number of feature maps in  base resolution.
In each new stage, the number of feature maps is doubled.
For example, with $f=24$ the network has about 354K learnable parameters.
Independent from $f$, the network has a theoretical perceptive field of $49 \times 49 \times 49$ voxels, which was found to be suitable  in preliminary experiments.

All convolutional layers use $3 \times 3 \times 3$ kernels and a leaky rectified linear unit \cite{xu2020reluplex} as the activation function, except for the very last layer, where the sigmoid function is applied instead.
Applying the sigmoid function to the output of the last convolution yields values in a range of (0, 1), which are interpreted as the probability of the respective voxel to be occupied by the surface.
The downsampling and upsampling operations are performed by nearest neighbour interpolation along all three spatial dimensions with scaling factors of 0.5 and 2.0, respectively.

\renewcommand{\arraystretch}{1.2}
\begin{center}
    \begin{table} 
    \centering
        \caption{\label{tab:arch}
    Layers of the architecture of $\mathcal{R}$.
    3D-Conv: 3D Convolutional layer.
    LRL: Leaky ReLU.
    BN: 2D batch normalization;
    Cat($L_X$): Depth-wise concatenation of the output of layer \textit{$L_X$} and the current layer. side-length: Output dimensions. $w$ is the side-length of the input shell $S_i$.
    }
    \begin{tabular}{c l  c  c  c }
            &Layer name & Type & side-length & num. chn. \\
        \hline
        \hline
        \multirow{7}{*}{\rotatebox[origin=c]{90}{\parbox[c]{1.5cm}{\centering Encoder}}}
        &   &  Input layer      & $n$    &    1    \\
        &Conv-(1,2) &  3D-Conv, LRL     & $n$    &    $f$  \\
        &Dw-1   &  Downsample       & $n/2$  &    $f$  \\
        &Conv-(3,4) &  3D-Conv, LRL     & $n/2$  &    $2f$ \\
        &Dw-2   &  Downsample       & $n/4$  &    $2f$  \\
        &Conv-(5,6) &  3D-Conv, LRL     & $n/4$  &    $4f$ \\
        &Conv-7  &  3D-Conv, LRL     & $n/4$  &    $2f$ \\
        \hline
        \multirow{8}{*}{\rotatebox[origin=c]{90}{\parbox[c]{2.7cm}{\centering Decoder}}}
        &Up-1  &  Upsample, Cat(6) & $n/2$  &    $(2+2)f$  \\
        &Conv-8  &  3D-Conv, LRL     & $n/2$  &    $2f$ \\
        &Conv-9  &  3D-Conv, LRL     & $n/2$  &    $f$ \\
        &Up-2  &  Upsample, Cat(3) & $n$    &    $(1+1)f$  \\
        &Conv-10  &  3D-Conv, LRL     & $n$    &    $f$ \\
        &Conv-11  &  3D-Conv, Sigmoid & $n$    &    $1$  \\
    \end{tabular}\\
\end{table}
\end{center}

\subsection{Loss Function and Training}\label{sec:training}

The loss function used in this work is a variant of the dice-loss, which is applied  in classification problems \citep{Sudre.etal2017}. 
This loss is well suited for our problem because it insusceptible to data imbalance. 
In our case, such imbalance occurs since there is much more empty space than occupied voxels in the reconstruction task. 

Let $V_i^\beta$ be a binary version of a voxel grid $V_i$, where 

\begin{equation*}\label{eq:binarization}
V_{i,(\hat{x},\hat{y},\hat{z})}^\beta = \begin{cases}
1 & \text{if } V_{i,(\hat{x},\hat{y},\hat{z})} \leq t_b, \\
0 & otherwise.
\end{cases}
\end{equation*}

Here, $t_b$ is a threshold value for the minimum number of 3D points in each voxel cell so it will be considered occupied.
In our experiments, we set $t_b = 2$. 
We assume that voxels which contain only a single 3D point are more likely to represent noise. Therefore, this is a measure  aimed to deal with noise, so that only voxels with more than two points are regarded. 

Another way to accommodate for noise in the data is by introducing a midified verison of the dice-loss. This version  uses weights at voxel level.
As voxels that hold only one point may distract the regression model, they should be ignored.
To this end, a weight tensor $W_i$ is computed for each voxel grid $V_i$, where $W_{i,(x,y,z)} = 0$ if $V_{i,(x,y,z)} = 1$ and  $W_{i,(x,y,z)} = 1$, otherwise. 

The basic reconstruction error is 

\begin{equation}
\label{eq:reconstruction-error}
    \mathcal{R}(\hat{V}_i,V_i^\beta) =1 - \frac{\mathcal{I}(\hat{V}_i, V_i^\beta)}{\mathcal{U}(\hat{V}_i,V_i^\beta)}
\end{equation}

with the intersection term, $ \mathcal{I} $,

\begin{equation}
    \mathcal{I}(\hat{V}_i,V_i^\beta) =  \sum_{\hat{x}=0}^{n-1} \sum_{\hat{y}=0}^{n-1} \sum_{\hat{z}=0}^{n-1} \hat{V}_{i,(\hat{x},\hat{y},\hat{z})} \cdot V_{i,(\hat{x},\hat{y},\hat{z})}^\beta
\end{equation}

and the weighted union term, $ \mathcal{U} $,

\begin{equation}
    \mathcal{U}(\hat{V}_i,V_i^\beta) =  \sum_{\hat{x}=0}^{n-1} \sum_{\hat{y}=0}^{n-1} \sum_{\hat{z}=0}^{n-1} \max(\hat{V}_{i,(\hat{x},\hat{y},\hat{z})}, V_{i,(\hat{x},\hat{y},\hat{z})}^\beta)\cdot W_{i,(\hat{x},\hat{y},\hat{z})}.
\end{equation}

Using this formulation of the reconstruction error,
the overall training loss $\mathcal{L}$ is

\begin{equation}
    \mathcal{L} = \dfrac{1}{B} \sum_{b=0}^B  \mathcal{R}(\hat{V}_b, V_b^\beta)
\end{equation}

where $B$ is the batch size and $b$ is the index of a sample in the batch.
Note that in preliminary experiments, we found this loss to outperform other loss definitions.
Particularly, we compared to minimizing the mean squared error and a variant of the dice-loss without weighting.

To train the network, its parameters are randomly initialized and then iteratively updated using ADAM optimizer with a learning rate of $\lambda = 0.0001$ and hyper-parameters $\beta_1 = 0.0$ and $\beta_2 = 0.999$.
During training, we sample a batch of $B$ voxel grids ($V_i$) from the point cloud.
Data augmentation is performed by randomly rotating the point cloud along the height axis before extracting each voxel grid.
This is done after selecting a random point to be the centre of the voxel grid. The grid position is then frozen and the point cloud is randomly translated along the height axis before the voxilization step.
In preliminary experiments we found that this step improves   the trained models substantially, with respect to the reconstruction capability. 
Models that were trained without this augmentation step tended to be biased towards predicting occupied voxels in the centre of the voxel grid.
Following the data augmentation step, the voxel grids in the batch are binarized. 

Next, the shells ($S_i$) are created as described in \Sref{sec:method_general}.
These  are presented to the network resulting in a predicted voxel grid ($\hat{V}_i$) for each shell in the batch. 
Using Eq. \ref{eq:reconstruction-error} the reconstruction error for each sample is calculated. 
The average reconstruction error over all samples in a batch corresponds to the reconstruction loss of the batch.
The parameters of the network are then iteratively updated using ADAM optimizer to minimize the reconstruction loss.
Training is stopped when the performance on a validation subset does not increase for $n_{ST}$ iterations. 
The parameter set resulting in the highest validation performance is used for the inference.
The performance measures are described in \Sref{sec:saliency_ratio}.

\subsection{Inference}\label{sec:inference}

After training, $R$ is used to predict the voxel grid $\hat{V}_i$ for the extracted shell $S_i$ of each point $p_i \in P$.
The reconstruction error $\mathcal{R}(\hat{V}_i, V_i)$ is then interpreted as a measure of saliency for $p_i$. Eventually, a saliency map is received, where each point has a saliency score of
\begin{equation}
\label{eq:saliency-value}
	\xi_i = \mathcal{R}(\hat{V}_i, V_i).
\end{equation}

\section{Test setup}\label{EVALUATION}

\subsection{Experiment setup}

Experiments were carried out using  an AMD Ryzen Theadripper 1900X 8-core processor machine with a CPU memory of 32GB and an NVIDIA GeForce RTX 2080 Ti GPU. 

Network parameters were optimized according to \Sref{sec:training}.
In all datasets, the batch size $B$ was set to 16 voxel grids per batch. 
The classifier was evaluated on the validation sets every 1,000 training iterations.
The hyper-parameter $n_{ST}$, which is the training stop parameter, was set to 10,000 iterations. The shell size $ m $ was set to $ 3 $ for all datasets. 

\subsection{Datasets}
To demonstrate the proposed method we used three datasets that differ by scene, acquisition platform, extent, number of points, point spacing, etc. 
In the following, we characterize each dataset. \Tref{tab:datasets} provides a summary of the key characteristics. 

\begin{figure}
	\centering
	\includegraphics[width=.7\linewidth]{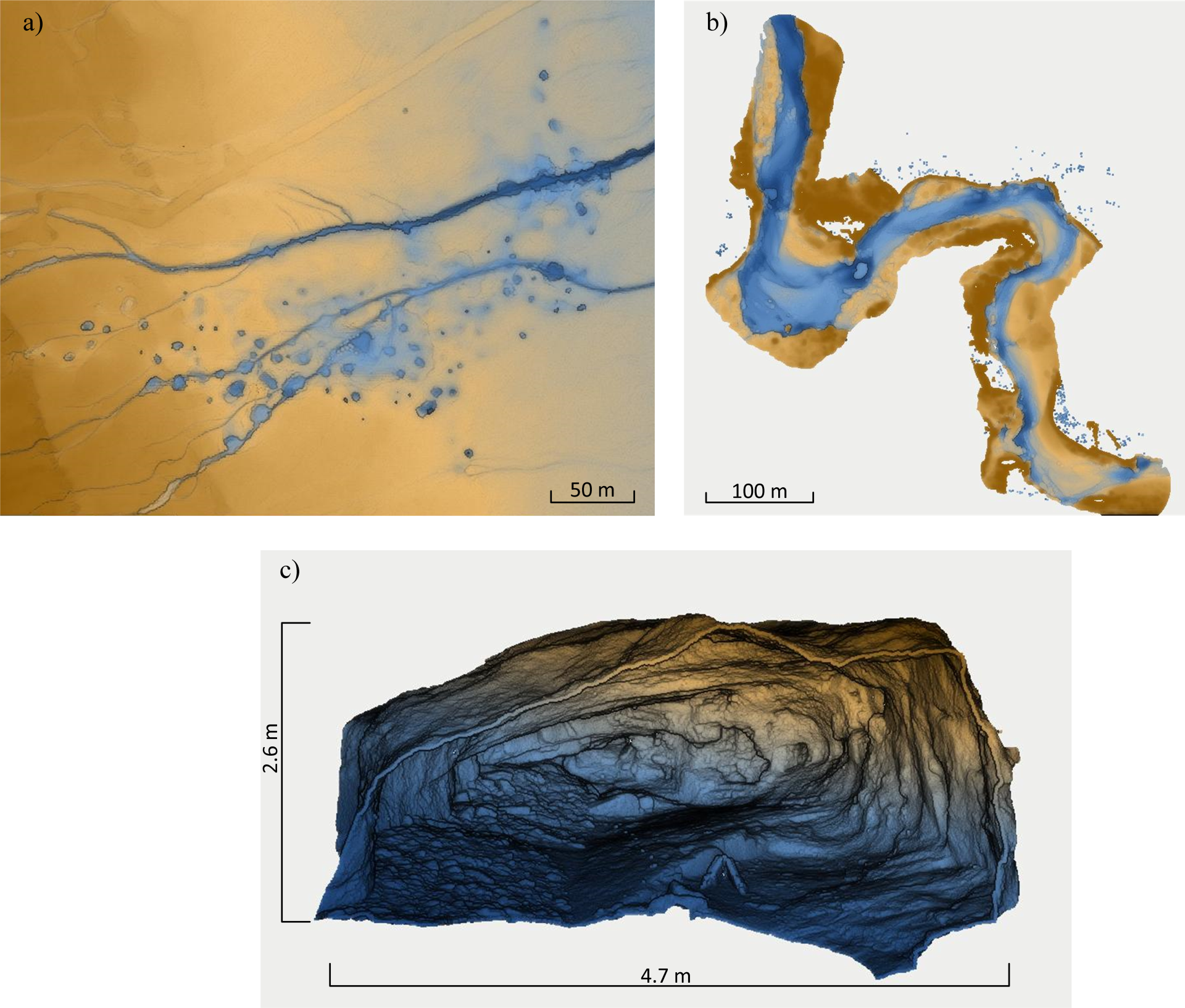}
	\caption{Datasets analysed in the study. a) Airborne dataset (Dead Sea Coast); b) UAV-borne dataset (Pielach River); c) Terrestrial dataset (Traisenbacher cave). Colours refer to elevation.}
	\label{fig:datasets}
\end{figure}

\begin{table}
   \centering
\begin{threeparttable}
	\begin{tabular}{lcccp{2cm}c}
	\centering
	
  \begin{minipage}[b]{1cm}\centering Dataset \# \end{minipage}& 
		\begin{minipage}[b]{1.5cm}\centering Scanning platform \end{minipage}& Scanner type & PRR$^\dagger$ [kHz] &
		\begin{minipage}[b]{1.97cm} \centering Mean point spacing [m]	\end{minipage}& No. of points\\ \hline \hline
		I& Airborne & Optech ALTM 2050 & {100} & \centering 0.5 & 1,632,928\\
		II & Airborne & Riegl VQ880-GH & 200& \centering 0.075&  50,813,569\\
		III & Terrestrial & Riegl VZ2000 & {550} & \centering 0.01& 786,267\\ \hline 
		
	\end{tabular}
	\caption{Acquisition characteristics of the analysed datasets.}\label{tab:datasets}
	\begin{tablenotes}
		\item [$\dagger$] Pulse repetition rate
	\end{tablenotes}
\end{threeparttable}
\end{table}

\begin{description}
	\item[Dataset \#I.] An airborne laser scan of an alluvial fan along the Dead Sea coast, Israel (open to the public  \citep{GSI.Arav2013}). It holds above 1.5 million points, at 0.5~m point spacing. The scanned surface is relatively flat, punctured by sinkholes and dissected by gullies \PFrefsub{fig:datasets}{a}. Being an airborne laser scan, some overlapping scanlines exist, which leads to a change in point density in some regions. 
	
	\item[Dataset \#II.] An airborne topo-bathymetric laser scan of a 750~m long section of a meandering river (Pielach River, Austria; \Frefsub{fig:datasets}{b}).  
    This scanner is characterised by its elliptic scanning pattern, which affects the average point density throughout the scan \citep{Arav.etal2024}.  
 This dataset holds over 50 million points. 
	Focusing on the river, vegetation was removed using the hierarchic robust interpolation method \citep{Pfeifer.Mandlburger2018} as implemented in OPALS \citep{Pfeifer.etal2014}.

	\item[Dataset \#III.] A terrestrial laser scan of  a small cave, the Untere Traisenbacher Höhle, Austria (\Frefsub{fig:datasets}{c}; open to the public \cite{Wimmer.Oberender2022}). Representing a cave, this dataset is fully three-dimensional, which makes it a challenging scene to analyse \citep{Arav.etal2022a}. 
	A single scanning position was used here. Therefore, on the one hand there are no ovelapping scanlines. On the other hand, the scan features occlusions, as there were no additional positions to mitigate them. These occlusions are characteristic to terrestrial laser scans in general, and in cave measurements in particular. Hence, this scene is a good example for 3D terrestrial scan.

\end{description}

\subsection{Validation and test subsets}\label{sec:subsets}
In each dataset, we specify two types of subsets: a validation subset ($D$) -- for stopping the training process and for tuning the hyper-parameters; a test subset ($T$) -- for testing and comparison purposes.
Each subset is divided into `salient' ($H$) and `non-salient' ($L$) regions. 
These correspond to the expected regions that should have higher and lower saliency scores, respectively. 

From each dataset a different number of subsets was extracted, depending on the scene.  
Non-salient areas were selected after visual inspection, to minimize the existence of salient regions within them. However,
 as   delineation was done manually, the subsets still included some small parts of the other class (i.e., `salient' in `non-salient' regions, and vice versa). 
 Nonetheless, the analysis only compares mean values of the same regions, so that inaccuracies in sampling are insignificant. 

Since saliency estimation is a subjective measure \citep{Akman.Jonker2010}, we describe below which objects/areas we expect to have higher saliency scores in each dataset.  
Accordingly, we define the minimal object size. The voxel size is then set to be  half of the minimal object size.
\Tref{tab:params} summarizes these features. 

\begin{description}
	\item[Dataset \#I.] Salient areas are defined either as sinkholes or as parts of gullies (e.g., \Frefsub{fig:zeelimsubsets}{a-b}). The sinkholes typically have 4-20~m diameter, while gullies are 2-9~m wide. 
     Therefore, the minimal size is 2~m and the consequent voxel size is set to 1~m \PTref{tab:params}. 
    As for the non-salient, these reflect the fan surface (\Frefsub{fig:zeelimsubsets}{c}). 
	A total of nine regions were extracted as salient areas and nine as non-salient ones. 
	
	\begin{figure}
		\centering
		\includegraphics[width=0.7\linewidth]{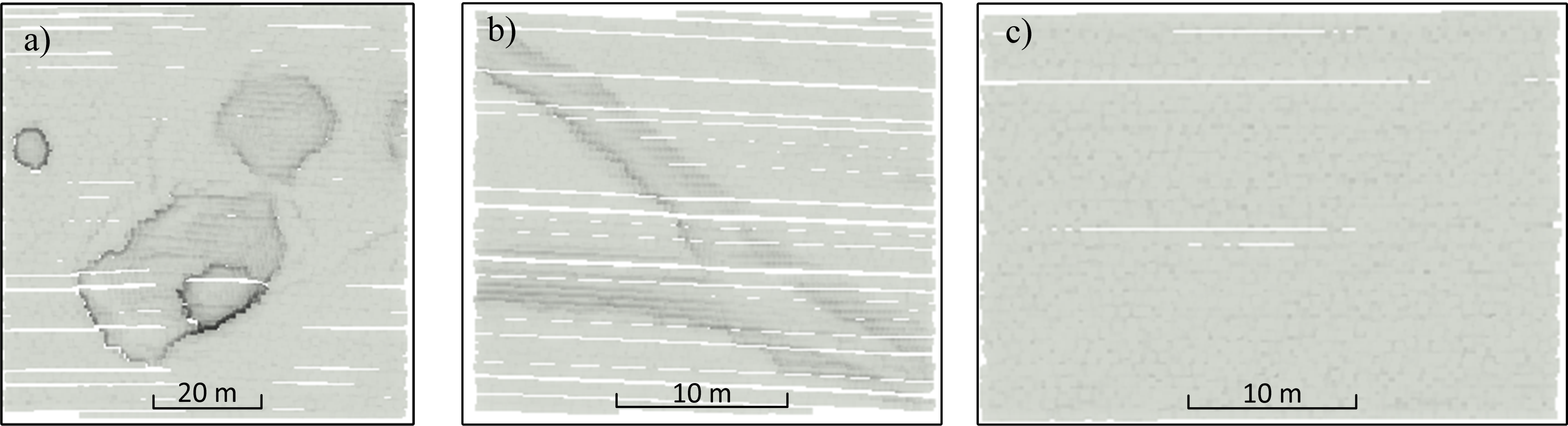}
		\caption{Examples of subsets chosen for validation in the dataset \#I. (a) and (b) - salient regions; (c) non-salient. Note the different scales.}
		\label{fig:zeelimsubsets}
	\end{figure}
	
	\item[Dataset \#II.] Salient features are defined as the riverbanks as well as objects on the riverbed that are larger than 0.3~m (e.g., driftwood, stone blocks). Accordingly, the voxel size is set to 0.15~m \PTref{tab:params}. 
  Three areas with  stone blocks, which were extracted in previous works  \citep{Mandlburger.etal2015}, were used as `salient' subsets   \PFref{fig:pielachsubsetssalient}.  
  Of these, two were chosen for testing and one for validation. The low number of extracted regions is a result of the complexity of the terrain. 
	Non-salient regions were chosen along the river and  reflect the riverbed which has varying surface roughness  \PFref{fig:pielachsubsetsnonsalient}. 
	
	\begin{figure}
		\centering
\begingroup%
  \makeatletter%
  \providecommand\color[2][]{%
    \errmessage{(Inkscape) Color is used for the text in Inkscape, but the package 'color.sty' is not loaded}%
    \renewcommand\color[2][]{}%
  }%
  \providecommand\transparent[1]{%
    \errmessage{(Inkscape) Transparency is used (non-zero) for the text in Inkscape, but the package 'transparent.sty' is not loaded}%
    \renewcommand\transparent[1]{}%
  }%
  \providecommand\rotatebox[2]{#2}%
  \newcommand*\fsize{\dimexpr\f@size pt\relax}%
  \newcommand*\lineheight[1]{\fontsize{\fsize}{#1\fsize}\selectfont}%
  \ifx\svgwidth\undefined%
    \setlength{\unitlength}{362.25bp}%
    \ifx\svgscale\undefined%
      \relax%
    \else%
      \setlength{\unitlength}{\unitlength * \real{\svgscale}}%
    \fi%
  \else%
    \setlength{\unitlength}{\svgwidth}%
  \fi%
  \global\let\svgwidth\undefined%
  \global\let\svgscale\undefined%
  \makeatother%
  \begin{picture}(1,0.96087423)%
    \lineheight{1}%
    \setlength\tabcolsep{0pt}%
    \put(0,0){\includegraphics[width=\unitlength,page=1]{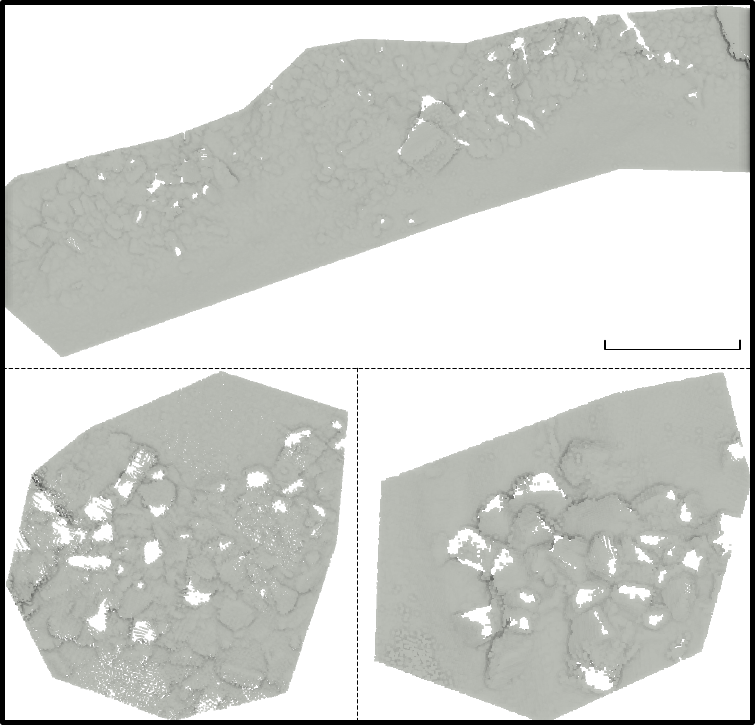}}%
    \put(0.85551559,0.50714036){\makebox(0,0)[lt]{\lineheight{1.25}\smash{\begin{tabular}[t]{l}10 m\end{tabular}}}}%
    \put(0,0){\includegraphics[width=\unitlength,page=2]{pielach_subsets_salient.pdf}}%
    \put(0.37063015,0.03049332){\makebox(0,0)[lt]{\lineheight{1.25}\smash{\begin{tabular}[t]{l}2.5 m\end{tabular}}}}%
    \put(0,0){\includegraphics[width=\unitlength,page=3]{pielach_subsets_salient.pdf}}%
    \put(0.87550016,0.02221174){\makebox(0,0)[lt]{\lineheight{1.25}\smash{\begin{tabular}[t]{l}2.5 m\end{tabular}}}}%
    \put(0.01283671,0.91614061){\makebox(0,0)[lt]{\lineheight{1.25}\smash{\begin{tabular}[t]{l}a)\end{tabular}}}}%
    \put(0.01229013,0.43180482){\makebox(0,0)[lt]{\lineheight{1.25}\smash{\begin{tabular}[t]{l}b)\end{tabular}}}}%
    \put(0.49379901,0.43215266){\makebox(0,0)[lt]{\lineheight{1.25}\smash{\begin{tabular}[t]{l}c)\end{tabular}}}}%
  \end{picture}%
\endgroup%

		\caption{Examples of high salient score subsets ($H$) chosen from the dataset \#II for a-b) testing; c) validation.  Note the different scales and region shapes.}
		\label{fig:pielachsubsetssalient}
	\end{figure}
	
	\begin{figure}
		\centering
		\includegraphics[width=\linewidth]{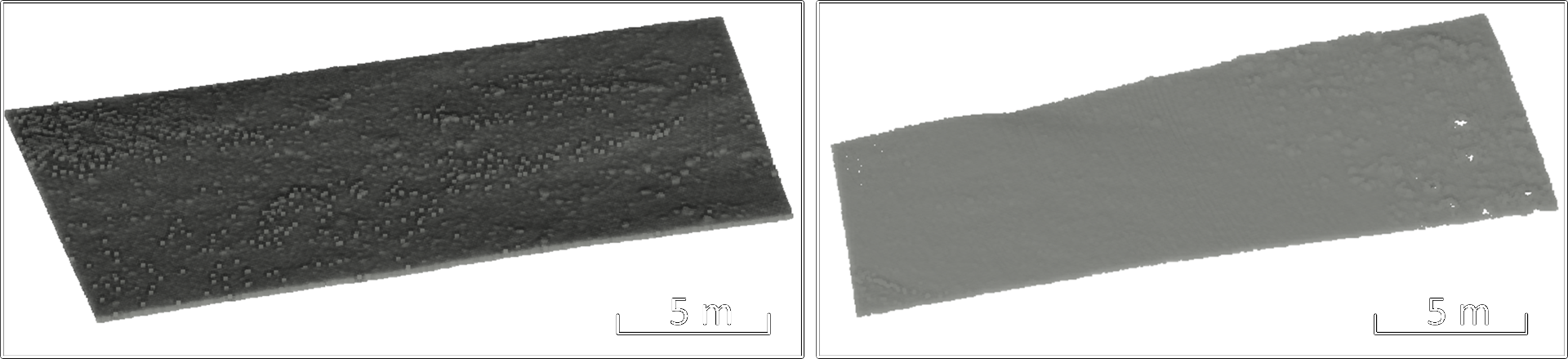}
		\caption{Dataset \#II. Examples of non salient subsets ($\nu$).  Note that the surface is not smooth but has some roughness (small stones).}
		\label{fig:pielachsubsetsnonsalient}
	\end{figure}
	
	\item[Dataset \#III.] Salient features refer to niches and pockets in the cave's walls and ceiling, as well as to some ledges and objects on the floor, with a minimal size of 0.1~m. Consequently, the voxel size was set to 0.05~m \PTref{tab:params}.
    To provide well-distributed subsets, both salient and non-salient subsets were chosen from the walls, the ceiling, and the floor. 
    While salient subsets were chosen to include niches and blocks \PFrefsub{fig:cavesubsets}{a}, non-salient subsets were focusing on the walls and ceiling that did not include any apparent niches \PFrefsub{fig:cavesubsets}{b}.
	
	\begin{figure}
		\centering
		\includegraphics[width=\linewidth]{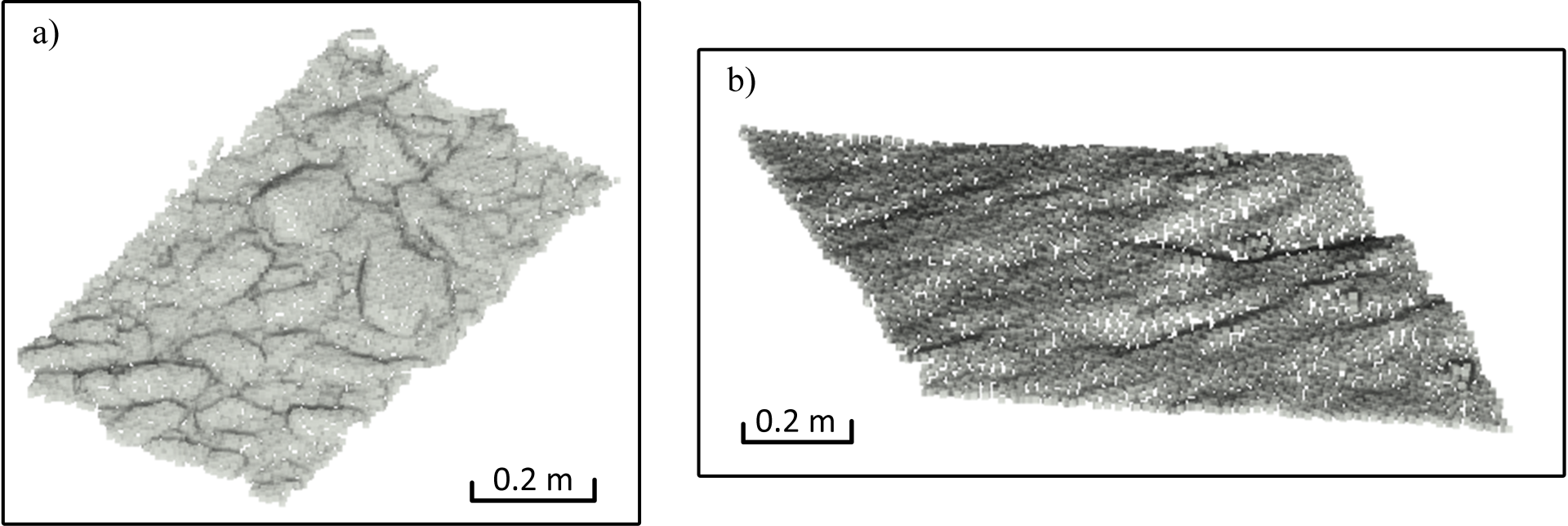}
		\caption{Examples of (a) salient and (b) non-salient subsets extracted from dataset \#III.  Note that the cave surface is fairly rough. }
		\label{fig:cavesubsets}
	\end{figure}
	
\end{description}

\begin{table}
	\centering
	
	\begin{tabular}{cp{2.7cm}cccccc}
		Dataset \# & 
		\begin{minipage}{2.5cm}\centering Predefined salient features \end{minipage}& 
		\begin{minipage}{1.9cm} \centering
			Min.\ object size [m]
		\end{minipage}&
		\begin{minipage}{1.9cm} \centering
			Voxel size  [m] 
		\end{minipage}& 
		\multicolumn{2}{c}{\begin{minipage}{1.8cm}
				\centering No.  of test sets
		\end{minipage}} & 
		\multicolumn{2}{c}{\begin{minipage}{2.5cm}
				\centering 		No. of validation sets 
		\end{minipage}}
		\\ 
		&&&  &$ H $& $ L $   & $ H $& $ L $\\ \hline \hline
		I& Gullies, sinkholes & 2 & 1 & 6   & 6 &3&3\\
		II & Riverbanks, stone blocks & 0.3 &0.2 & 2&3&1& 3  \\
		III & Boulders, niches, pockets &0.1& 0.05 & 4  &4&3&3\\ \hline 
		
	\end{tabular}
	\caption{Defined salient features in each dataset, minimal object and voxel sizes, as well as number of test and validation subsets.}\label{tab:params}
\end{table}

\subsection{Evaluation metrics}\label{sec:saliency_ratio}

Measuring the performance of saliency scores is difficult. This is because the success rate cannot be easily quantified and may depend on user's understanding of the data \citep{Hillier.etal2014, Scheiber.etal2015, Vinci.etal2016}. 
Moreover, since we use saliency as a relative measure within the dataset, it is impossible to compare values of one method to another. 
In most reviewed literature, saliency was used as a preliminary step for other analyses \citep[e.g.,][]{Laazoufi.Hassouni2022, Liang.etal2023}. Then, the quantitative quality was measured according to the success rate of the procedures that follow. For example, \citet{Tinchev.etal2021} assessed the registration quality, which was carried out based on keypoint detection using estimated saliency. Other works were comparing the results to existing benchmarks \citep[e.g.,][]{Fan.etal2022}. Such a benchmark does not exist in our case. 
Therefore, we propose a saliency ratio for quantitative evaluation in addition to the visual inspection of the results. 

Using the subsets $D$ and $T$ defined in \Sref{sec:subsets}, we define the saliency ratio $\hat{r}$.
In particular, we define the ratio
\begin{equation}\label{eq:ratio_validation}
 \hat{r}_D = \dfrac{\bar{\xi}_{D,H}}{\bar{\xi}_{D,L}}
\end{equation}
with  $\bar{\xi}_{D,H}$  the mean saliency score for points with a high expected saliency scores and $\bar{\xi}_{D,L}$ the mean saliency score for those with a low expected saliency, both in the validation subsets. 
Similarly, for the final testing of the method, we define 
\begin{equation}\label{eq:ratio_test}
 \hat{r}_T = \dfrac{\bar{\xi}_{T,H}}{\bar{\xi}_{T,L}}   
\end{equation}
using the test subsets ($T$) instead of the validation ($D$). 

Ratios that are larger than $ 1 $ suggest that the mean estimated saliency scores in $H$  is higher than those in $L$, which is the expected result. As these ratios approach 1, the difference in estimated scores between salient and non-salient regions decreases. That is to say, the distinction between the two regions decreases. When the ratio is smaller than 1, the saliency was not estimated correctly, as regions that are expected to be with lower values yielded higher ones, and vice versa.

The metric $\hat{r}_D$ is used to tune the hyper-parameters of the method.
The metric $\hat{r}_T$, which assesses the performance on the test subsets, is used to compare the method to existing approaches.

\subsection{Baseline approaches}\label{sec:baselines}
To compare our method to state-of-the-art, we used the following two baseline methods:

\subsubsection{Plane-based approach}\label{sec:plane}
Given that our methodology hinges on reconstructing topography surfaces that may exhibit local planarity, a planar reconstruction is worth examining. Therefore, to highlight the merits of our learning approach, we advocate for its comparison against a plane-based anomaly search. 
With this in mind, we propose an alternative strategy to reconstruct the core of a voxel grid $V_i$ by leveraging its shell $S_i$. 
At the heart of this method is the concept of fitting a plane to the voxels within the shell and subsequently projecting this plane onto the voxel grid, resulting in the reconstructed grid $\hat{V}_i$.

To find the best-fit plane for the shell voxels, we commence by computing the coordinates covariance matrix of each of the voxels that lie on the shell. 
The eigenvector corresponding to the smallest eigenvalue of this matrix provides us with the normal vector $\mathbf{n}$ of the optimal plane. 
Combined with the distance $d$ from the origin, this establishes the plane equation in 3D space.

For any voxel $(\hat{x}, \hat{y}, \hat{z})$ within this space, its perpendicular distance $d_{p_{\hat{x}, \hat{y}, \hat{z}}}$ from the plane is derived from the plane's equation as:
$d_{p_{\hat{x}, \hat{y}, \hat{z}}} = | \mathbf{n} \cdot (\hat{x}, \hat{y}, \hat{z}) - d |$.
Here, $\mathbf{n} \cdot (\hat{x}, \hat{y}, \hat{z})$ represents the dot product between the normal vector and the voxel.

To represent the plane in the voxel grid, any voxels where $|d_{p_{\hat{x}, \hat{y}, \hat{z}}}| < t_d$, with $t_d = 0.5 \cdot w$ (with $w$ the voxel side length)  are assigned a value of one. 
By processing each voxel in this manner, the resultant grid is the reconstructed voxel grid $\hat{V}_i$. Then, the loss is computed by \Eref{eq:reconstruction-error} and the saliency is estimated by \Eref{eq:saliency-value}.
This way, the plane-based reconstruction is in fact a simplified comparative to our primary approach.

The voxel grid sizes to which the plane was fitted were chosen according to the those used in the proposed method, i.e., $ n=16,\ 24 $ and 32.

\subsubsection{Handcrafted saliency estimation}\label{sec:handcrafted}
We use the handcrafted saliency proposed in \cite{Arav.Filin2022} as another baseline method. This is because, to the best of our knowledge,  it is the only point cloud based saliency estimation method that is attuned for natural environments.  
It is based upon the assumption that when dealing with topography distinctness would not be apparent in the immediate surroundings of a point. Therefore, it uses a weighting function that gives lower weights to nearby points and higher weights to more distant ones.  
To do so, the size of the surroundings and the minimal object size are set. Here, we set these according to \Tref{tab:params}, where the voxel size corresponds to the size of the surroundings. 
The saliency is then evaluated according to the deviation in surface normals and curvature within the defined surrounding. It is calculated as 

\begin{equation}\label{eq:saliency}
	\xi_i=2 - \left[\exp\left(-d\textbf{n}(p_i)\right) +  \exp\left({-d\kappa}(p_i)\right)\right]
\end{equation}

with $d\textbf{n}$ and $d\kappa$ are the sum of deviations in normal and curvature within the defined surroundings. 

\subsection{Experiments description}
For each dataset, we first performed a tuning for two hyper-parameters: the number of features in base resolution, $ f $, and the voxel grid side length, $ n $. We focused on these two parameters as  they are considered the most important parameters of the method. 
We test their effect and discuss the saliency evaluation results achieved when using  different combinations of the two. In all experiments, we used  magnitudes of 8, 16, and 32 for $f$,  and 16, 24, and 32 for $n$.
The network was trained five times in each combination. 
After training, saliency scores were evaluated  for the validation subsets. The saliency ratio  $ \hat{r}_V $ (\Sref{sec:saliency_ratio}) was then computed. Eventually, the mean saliency ratio and its standard deviation over the five runs were evaluated. 
Then, based on the best achieved results, we evaluated saliency scores  for the entire dataset. 

For each dataset,  saliency scores  were also evaluated using the baseline methods  \PSref{sec:baselines}.  
The comparison is carried out by evaluating the saliency ratio for the test subsets $T$ for all applied methods. 
These, together with the visual impression of the saliency maps of the entire scene,  enabled an evaluation of the saliency results.

\section{Results and discussion}\label{RESULTS}


\subsection{Hyper-parameters tuning}\label{sec:tuning}
Tables \ref{tab:zeelim_rv}-\ref{tab:cave_rv} present the average saliency ratio results over five tests at each combination and for  datasets \#I, II, and III, respectively. 
It can be seen that in each  dataset, the saliency ratios using the different hyper-parameters are similar. 
These range between 2.2-2.5 in dataset \#I; 2.36-2.52 in dataset \#II; and 1.12-1.19 in dataset \#III (Tables \ref{tab:zeelim_rv}, \ref{tab:pielach_rv}, and \ref{tab:cave_rv}, respectively). 
Additionally, it can be seen that in all three datasets and for all combinations of $ f $ and $ n $ the saliency ratio is larger than 1. This indicates that  regions which are defined as salient have  higher saliency scores than the non-salient ones, as expected. 
However, a homoscedastic t-test did not show statistical  distinction between `salient' and `non-salient' at  85\% probability for the validation subsets.

\begin{table}
	\centering
	
	\renewcommand{\arraystretch}{1.2}
	\begin{tabular}{c|ccc }
		\backslashbox{$ n $}{$ f $} &  8  & 16  & 32   \\ 
		\hline \hline
		16	&\textbf{2.51 $\pm 0.08$}	&2.31$\pm 0.04$	&2.33$\pm 0.03$\\
		24	&2.35$\pm 0.04$	&2.29$\pm 0.03$	&2.30$\pm 0.01$ \\
		32	&2.20$\pm 0.03$	&2.21$\pm 0.01$	&2.24$\pm 0.03$\\
		\hline
	\end{tabular}
	\caption{Dataset \#I. Saliency ratio (average and standard deviation over 5 runs) for the validation subsets ($\hat{r}_D$). It can be seen that for all combinations, saliency scores are higher at salient regions than non-salient ones.  This implies that the proposed method highlighted salient regions correctly.}
	\label{tab:zeelim_rv}
	
\end{table}

\begin{table}
	\centering
	\renewcommand{\arraystretch}{1.2}
	\begin{tabular}{c|ccc }
		\backslashbox{$ n $}{$ f $} &  8  & 16  & 32   \\ 
		\hline \hline
		16 & $2.42\pm0.03$  & $2.40\pm0.01$ &  $2.41\pm0.03$ \\
		24 & $2.48\pm0.02$ & $2.49\pm0.02$  &  ${\bf 2.52\pm0.08}$ \\
		32 & $2.36\pm0.01$ & $2.39\pm0.02$  & $2.47\pm0.05$ \\ \hline
	\end{tabular}
	\caption{Dataset \#II. Saliency ratio (average and standard deviation over 5 runs) for the validation subsets ($\hat{r}_D$). It can be seen that for all combinations, saliency scores are higher at salient regions than non-salient ones. This implies that the proposed method highlighted salient regions correctly.}
	\label{tab:pielach_rv}
\end{table}

\begin{table}
	\centering
	\renewcommand{\arraystretch}{1.2}
	\begin{tabular}{c|ccc }
		\backslashbox{$ n $}{$ f $} &  8  & 16  & 32 
		\\ 
		\hline \hline
		16 & $1.12\pm0.04$  & $1.18\pm0.02$ &  $\bf{1.19\pm0.01}$ 
		\\
		24 & $1.15\pm0.02$ & $1.16\pm0.03$  &  $ 1.18\pm0.02$
		\\
		32 & $1.14\pm0.04$ & $1.16\pm0.01$  & $\bf{1.19\pm0.02}$
		\\ \hline
	\end{tabular}
	\caption{Dataset \#III. Saliency ratio (average and standard deviation over 5 runs) for the validation subsets ($\hat{r}_D$). For all combinations the ratio values are close to 1, implying that the difference between estimated salient and non-salient values is small. Despite that, the estimated ratios are still larger than 1, meaning that the method evaluated salient regions correctly.}
	\label{tab:cave_rv}
\end{table}

\begin{figure}
 \def\svgwidth{\textwidth}
  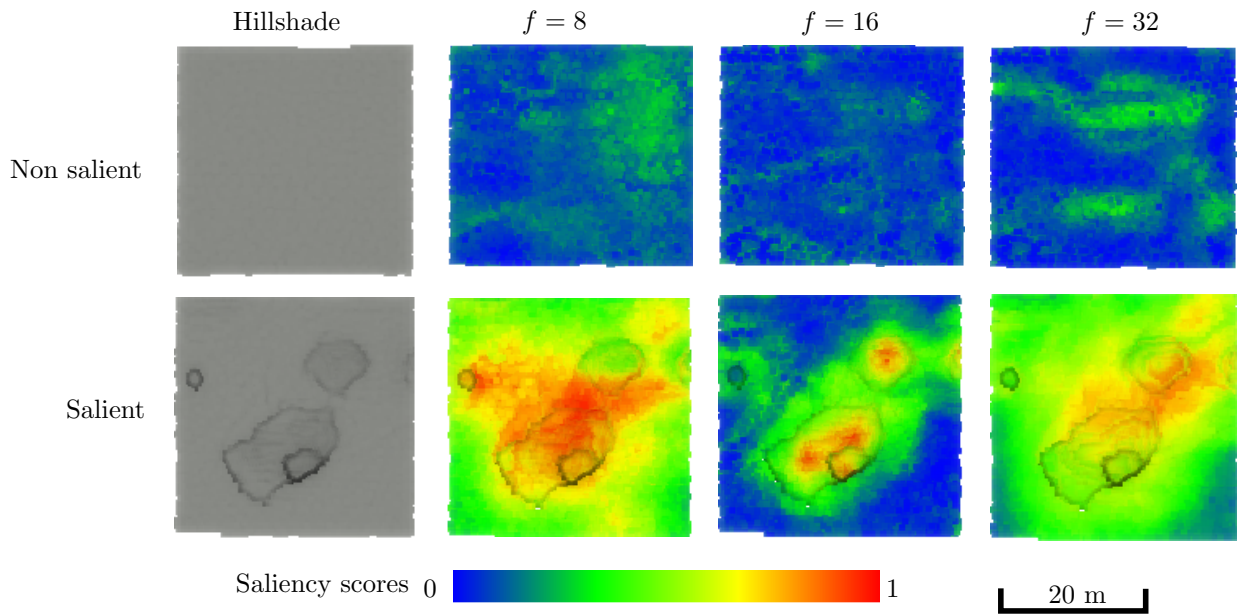
  \caption{Dataset \#I. Saliency scores estimated for the validation subsets with $ n=16 $ and different numbers of feature maps in base resolution. }
  	\label{fig:zeelimvaef}
\end{figure}

To better understand the effect of each hyper-parameter on the saliency map, we visually examine the results achieved when one parameter is fixed and the other changes. 
We begin by testing the effect of the number of feature maps in base resolution ($f$). To do so, we fixed the size of the voxel grid $n$ at the size which yielded the highest $\hat{r}_D$. 
\Fref{fig:zeelimvaef}  shows the results for $n=16$  using the validation subsets for dataset \#I. 
The effect of $f$ is mostly seen in the non-salient regions. 
There,  the least regions are being marked with high saliency scores when  $f=16$. This is because the number of features dictates the capacity of the network to reconstruct the surface.
Too few features in base resolution will lead to a larger discrepancy from the original point cloud, and thus to higher saliency scores in non-salient regions (e.g., $f=8$). 
On the other hand, too many features will lead to overfitting. Then, the reconstructed surface will deviate from the original cloud and result in incorrectly estimated high saliency scores  ($f=32$ in both \Fref{fig:zeelimvaef}). 
The number of features, however, may differ from one dataset to another, depending on the scene's surface. Therefore, it has to be tested for each dataset individually. 

\begin{figure}
	\centering
	\includegraphics[width=\textwidth]{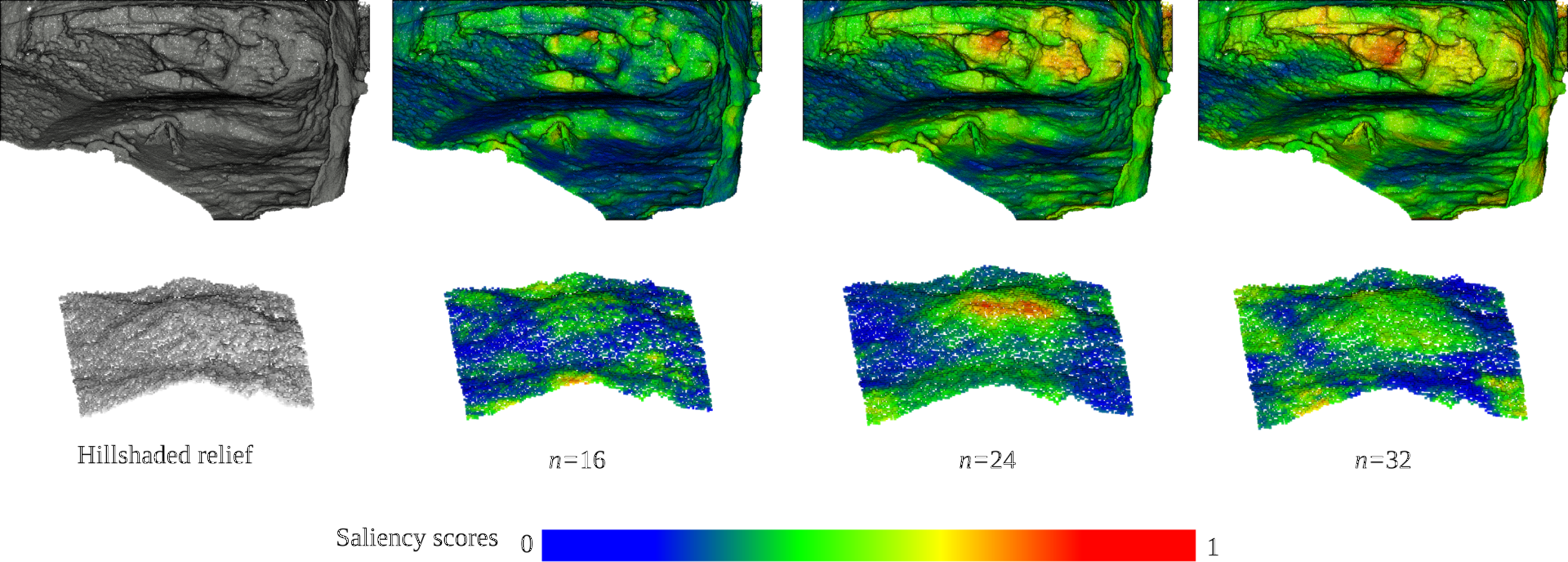}
	\caption{Dataset \#III. Saliency scores for $f=32$ at different sizes of the voxel grid. 
		It can be seen that as the grid size grows, more regions are marked as salient.}
	\label{fig:cavedetail}
\end{figure}

Similarly, we examined the effect of the voxel grid size, $n$, by fixing $f$ with the number that achieved the highest ratio. 
\Fref{fig:cavedetail} presents an example of  parts from the validation subset in dataset \#III,  where $f=32$ and $n= 16, 24,$ and 32.  
It shows  that it mainly affects the extent of the regions that receive higher saliency scores.  
The larger $n$ is, the larger the inferred area, and thus the discrepancy from the original point cloud is larger, leading to less localized marking. Therefore, as the grid size increases the highlighted area increases as well. 

It should be noted, however, that as the saliency ratio suggests, 
there are hardly any visual differences between the saliency maps generated by different hyper-parameters.

Between datasets, it can be seen that while dataset \#I yielded the largest $\hat{r}_D$, dataset \#III yielded the lowest. This fact can be attributed to the complexity of the analysed surfaces. 
In dataset \#I the terrain is quite smooth and almost planar;  dataset \#II features a rougher but still mostly planar terrain; and dataset \#III is composed of non-planar and mostly uneven and rough surfaces. 
This means that as the surface becomes less smooth (i.e., with higher surface variability), the network's ability to reconstruct the surface decreases, and thus the difference between  $\bar{\xi}_H$  and $\bar{\xi}_L$ decreases. 



\subsection{Saliency estimation}
The hyper-parameters used for the saliency evaluation for the entire datasets were those that produced the highest saliency ratio in the tuning phase \PSref{sec:tuning}. 
This is based on the assumption that these hyper-parameters will provide the most pronounced distinction between `salient' and `non-salient' regions. 
After the inference phase,  saliency ratios were evaluated for the test subsets (i.e., $\hat{r}_T$). These produced similar magnitudes as those estimated for the validation subsets \PTref{tab:rt_comparison}.
In the following, we present the saliency map of each dataset and discuss the results separately, as we compare them to the results of the baseline methods. 

\begin{table}
	\centering
	\renewcommand{\arraystretch}{1.2}
	\begin{tabular}{cccccc }
		 &$f$ & $n$&  Ours  & Plane-based  & Handcrafted 
		\\ 
		\hline \hline
		Dataset \#I 	&8 & 16 &	2.44	&1.43	&2.85 \\
		Dataset \#II	&32 & 24&2.49	&1.9	&1.07 \\
		Dataset \#III	&32 & 16 &1.23	&1.06	&10.9 
		\\ \hline
	\end{tabular}
	\caption{Mean saliency ratio values on test subsets using the hyper-parameters that yielded the largest saliency ratio values in the tuning phase. }
	\label{tab:rt_comparison}
\end{table}

\subsubsection{Dataset \#I}
\begin{figure}
	\centering
		   \def\svgwidth{\textwidth}
\begingroup%
  \makeatletter%
  \providecommand\color[2][]{%
    \errmessage{(Inkscape) Color is used for the text in Inkscape, but the package 'color.sty' is not loaded}%
    \renewcommand\color[2][]{}%
  }%
  \providecommand\transparent[1]{%
    \errmessage{(Inkscape) Transparency is used (non-zero) for the text in Inkscape, but the package 'transparent.sty' is not loaded}%
    \renewcommand\transparent[1]{}%
  }%
  \providecommand\rotatebox[2]{#2}%
  \newcommand*\fsize{\dimexpr\f@size pt\relax}%
  \newcommand*\lineheight[1]{\fontsize{\fsize}{#1\fsize}\selectfont}%
  \ifx\svgwidth\undefined%
    \setlength{\unitlength}{557.60418701bp}%
    \ifx\svgscale\undefined%
      \relax%
    \else%
      \setlength{\unitlength}{\unitlength * \real{\svgscale}}%
    \fi%
  \else%
    \setlength{\unitlength}{\svgwidth}%
  \fi%
  \global\let\svgwidth\undefined%
  \global\let\svgscale\undefined%
  \makeatother%
  \begin{picture}(1,0.85302427)%
    \lineheight{1}%
    \setlength\tabcolsep{0pt}%
    \put(0,0){\includegraphics[width=\unitlength,page=1]{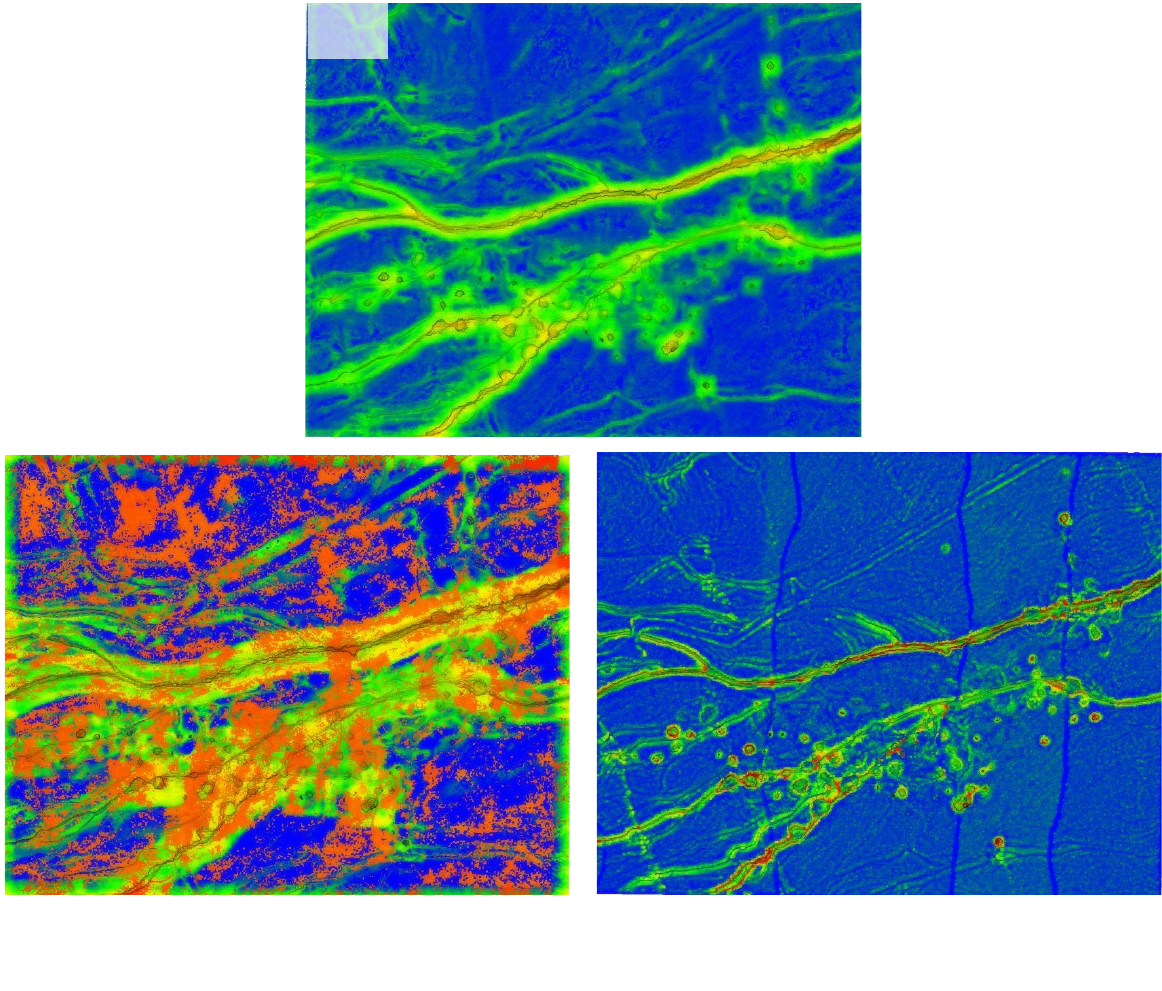}}%
    \put(0.27195973,0.82773321){\makebox(0,0)[lt]{\lineheight{1.25}\smash{\begin{tabular}[t]{l}a)\end{tabular}}}}%
    \put(0,0){\includegraphics[width=\unitlength,page=2]{zeelim_compare.pdf}}%
    \put(0.00717355,0.43601809){\makebox(0,0)[lt]{\lineheight{1.25}\smash{\begin{tabular}[t]{l}b)\end{tabular}}}}%
    \put(0,0){\includegraphics[width=\unitlength,page=3]{zeelim_compare.pdf}}%
    \put(0.26442306,0.00021016){\makebox(0,0)[lt]{\lineheight{1.25}\smash{\begin{tabular}[t]{l}0\end{tabular}}}}%
    \put(0.79110437,0.00021016){\makebox(0,0)[lt]{\lineheight{1.25}\smash{\begin{tabular}[t]{l}1\end{tabular}}}}%
    \put(0.12409663,0.03707831){\makebox(0,0)[lt]{\lineheight{1.25}\smash{\begin{tabular}[t]{l}Saliency scores\end{tabular}}}}%
    \put(0,0){\includegraphics[width=\unitlength,page=4]{zeelim_compare.pdf}}%
    \put(0.5209724,0.43750192){\makebox(0,0)[lt]{\lineheight{1.25}\smash{\begin{tabular}[t]{l}c)\end{tabular}}}}%
  \end{picture}%
\endgroup%

	\caption{Dataset \#I. Saliency results using a) proposed method with $f=8$ and $n=16$;
		b) Plane-based highlighting with $ n=16 $; 
		c) Handcrafted approach \citep{Arav.Filin2022} using $ \rho= 2$. }
	\label{fig:zeelimcompare}
\end{figure}

\Frefsub{fig:zeelimcompare}{a} shows the saliency map generated by the proposed approach for the dataset \#I using $f=8$ and $n=16$. 
It can be seen that the expected gullies and sinkholes were highlighted. 
Higher saliency scores were given for the gullies' thalweg, the bottom of the sinkholes, and to smaller channels.

\Frefsub{fig:zeelimcompare}{b and (c)} show the saliency maps generated by the baseline methods. 
We use $n=16$ for the plane-based method  and a minimal object size of 2~m for the handcrafted one. 
It can be seen that the plane-based method (b) yielded poor saliency map. Though the gullies did receive higher scores, these are lower than other regions that locally deviate from planarity. 
Furthermore, points that belong to sinkholes were not marked relative to their surroundings. Instead, they were grouped together with other highlighted regions.   
The handcrafted method provided a better picture \PFrefsub{fig:zeelimcompare}{c}. There, most gullies and sinkholes were highlighted as well as small micro-channels.  Still, the map seems noisy and regions with overlapping scanlines are marked as more salient (light green bounded by light blue). 
When comparing to the proposed method, the impression of the saliency map is of more consistent salient regions and less noise. 

\Tref{tab:rt_comparison} shows the saliency ratio evaluation of the test subsets for each method.  
The plane-based approach shows the smallest difference, with a ratio of 1.43. 
The handcrafted approach yielded the highest ratio of $  \hat{r}_T=2.85 $. This is in the same scale of the proposed method ($\hat{r}_T=2.44$). It is important to mention that  no statistical significance was found between $H$ and $L$ testing subsets using a homoscedastic t-test for all saliency methods (at 85\% probability). 

These results are substantiated by the visual map of the detected saliency \PFrefsub{fig:zeelimcompare}{b}. 
 
\subsubsection{Dataset \#II}
\begin{figure}
	\centering
		   \def\svgwidth{\textwidth}
	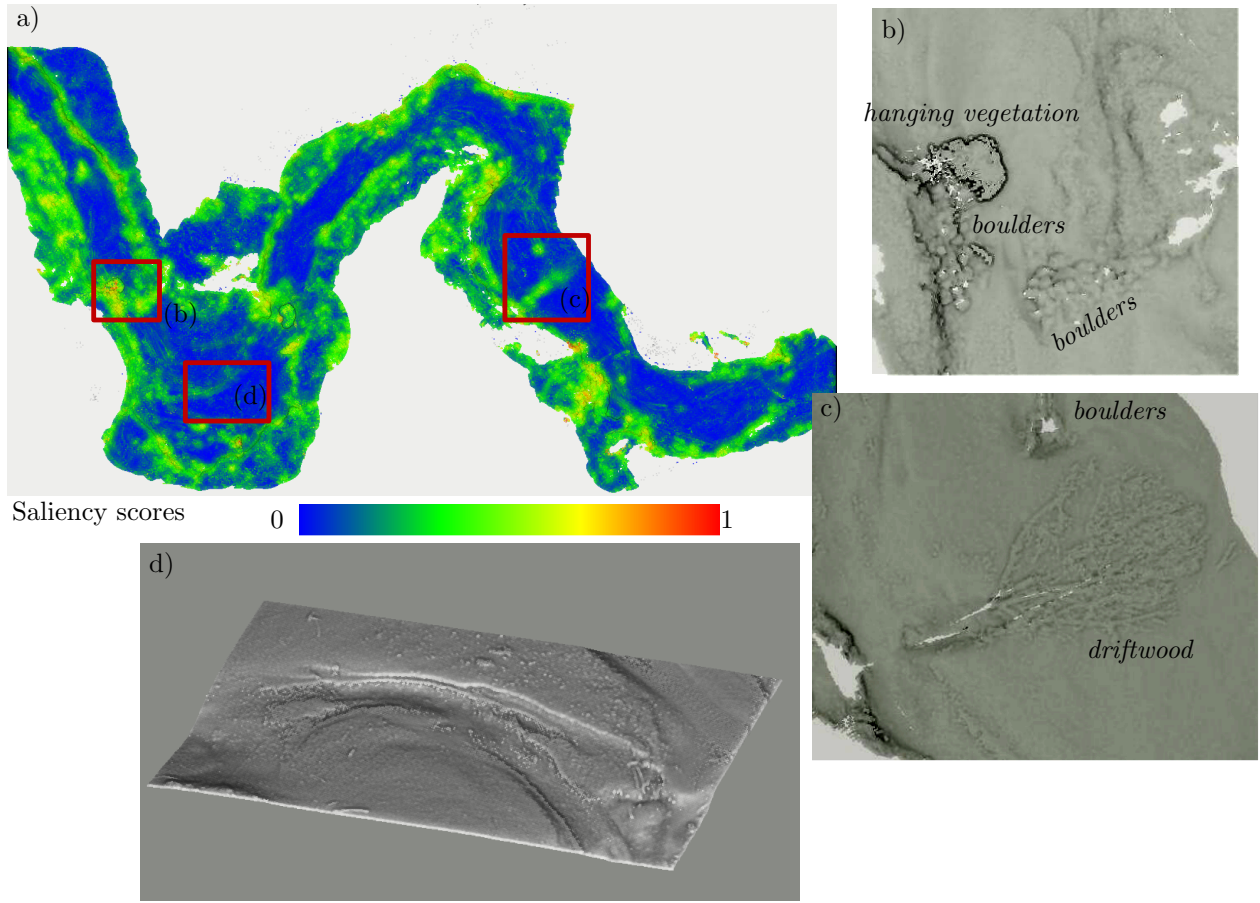
	\caption{Top: (a) saliency in dataset \#II using 0.2~m voxel size, $n=24$ and $f=8$. (b-d) Hillshaded representation of regions on the riverbed that were detected as salient -- (b) submerged boulders and vegetation; (c) submerged driftwood and boulders; and (c) banks of a small gully that was incised within the riverbed.}
	\label{fig:pielachvae}
\end{figure}

\Frefsub{fig:pielachvae}{a} shows the saliency scores using $f=32$ and $n=24$. It can be seen that  higher saliency scores refer to the river banks. 
However, a closer inspection discovers other expected features on the riverbed, such as boulders and hanging vegetation \PFrefsub{fig:pielachvae}{b}. Notably, other entities were found, e.g., submerged driftwood and a small incised gully \PFrefsub{fig:pielachvae}{c-d}. This result emphasizes the advantages of the proposed method: the searched features are not defined in advance, only the minimal size of interesting features needs to be specified. 

\begin{figure}
	\centering
		   \def\svgwidth{\textwidth}
\begingroup%
  \makeatletter%
  \providecommand\color[2][]{%
    \errmessage{(Inkscape) Color is used for the text in Inkscape, but the package 'color.sty' is not loaded}%
    \renewcommand\color[2][]{}%
  }%
  \providecommand\transparent[1]{%
    \errmessage{(Inkscape) Transparency is used (non-zero) for the text in Inkscape, but the package 'transparent.sty' is not loaded}%
    \renewcommand\transparent[1]{}%
  }%
  \providecommand\rotatebox[2]{#2}%
  \newcommand*\fsize{\dimexpr\f@size pt\relax}%
  \newcommand*\lineheight[1]{\fontsize{\fsize}{#1\fsize}\selectfont}%
  \ifx\svgwidth\undefined%
    \setlength{\unitlength}{500.18447876bp}%
    \ifx\svgscale\undefined%
      \relax%
    \else%
      \setlength{\unitlength}{\unitlength * \real{\svgscale}}%
    \fi%
  \else%
    \setlength{\unitlength}{\svgwidth}%
  \fi%
  \global\let\svgwidth\undefined%
  \global\let\svgscale\undefined%
  \makeatother%
  \begin{picture}(1,1.12442913)%
    \lineheight{1}%
    \setlength\tabcolsep{0pt}%
    \put(0.02893463,0.8802503){\makebox(0,0)[lt]{\lineheight{1.25}\smash{\begin{tabular}[t]{l}(a)\end{tabular}}}}%
    \put(0.02893463,0.5507376){\makebox(0,0)[lt]{\lineheight{1.25}\smash{\begin{tabular}[t]{l}(b)\end{tabular}}}}%
    \put(0.02893463,0.22000245){\makebox(0,0)[lt]{\lineheight{1.25}\smash{\begin{tabular}[t]{l}(c)\end{tabular}}}}%
    \put(0,0){\includegraphics[width=\unitlength,page=1]{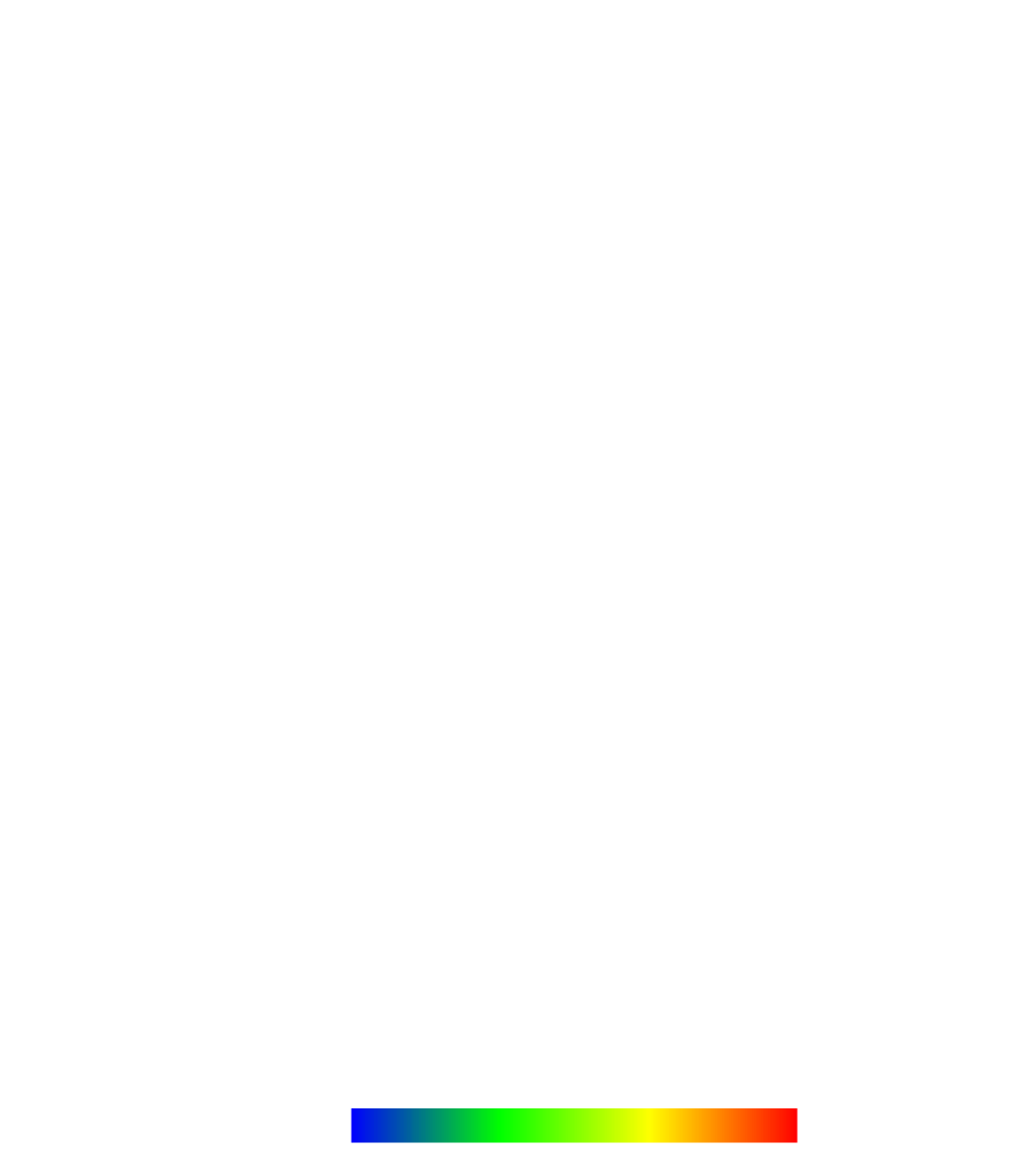}}%
    \put(0.32,0.03){\makebox(0,0)[lt]{\lineheight{1.25}\smash{\begin{tabular}[t]{l}0\end{tabular}}}}%
    \put(0.78,0.03){\makebox(0,0)[lt]{\lineheight{1.25}\smash{\begin{tabular}[t]{l}1\end{tabular}}}}%
    \put(0.15,0.03000041){\makebox(0,0)[lt]{\lineheight{1.25}\smash{\begin{tabular}[t]{l}Saliency scores\end{tabular}}}}%
    \put(0,0){\includegraphics[width=\unitlength,page=2]{pielach_compare.pdf}}%
    \put(0.18,1.0494568){\makebox(0,0)[lt]{\lineheight{1.25}\smash{\begin{tabular}[t]{l}Salient\end{tabular}}}}%
    \put(0.6,1.04856112){\makebox(0,0)[lt]{\lineheight{1.25}\smash{\begin{tabular}[t]{l}Non-salient\end{tabular}}}}%
  \end{picture}%
\endgroup%

	\caption{Dataset \#II. Saliency scores using the three methods on test regions: (a) proposed method using $n=24$ and $f=8$; 
	(b) plane-based reconstruction using $f=32$; 
	(c) handcrafted method using $\rho =0.25$~m.}
	\label{fig:pielachcompare}
\end{figure}

We use $f=32$ for the plane-based baseline method, and a minimal object size of $0.25$~m  for the  handcrafted baseline method. 
\Fref{fig:pielachcompare} presents the saliency results of the three methods in salient and non-salient test regions. 
Of the three methods, the best visual results were achieved for the proposed approach (a). There,  boulders are highlighted in the salient subset, whereas in the non-salient region, only the frame of the subset was marked as salient. This is an expected result, as it is more difficult to predict the surface at the edges, due to the lack of information and training data in these regions.  
The plane-based method highlighted most of the surface, irrespective to the data (b); 
the handcrafted method successfully highlighted some of the boulders (c, left), but  arbitrary patterns are marked in the non-salient subset (c, right). This maybe as a result of either the scanning  pattern, which yields overlapping scanlines, or due to the high surface roughness in this dataset. 
 The visual results are generally corroborated by the saliency ratios \PTref{tab:rt_comparison}. 


\subsubsection{Dataset \#III}
\begin{figure}
	\centering
	\includegraphics[width=\textwidth]{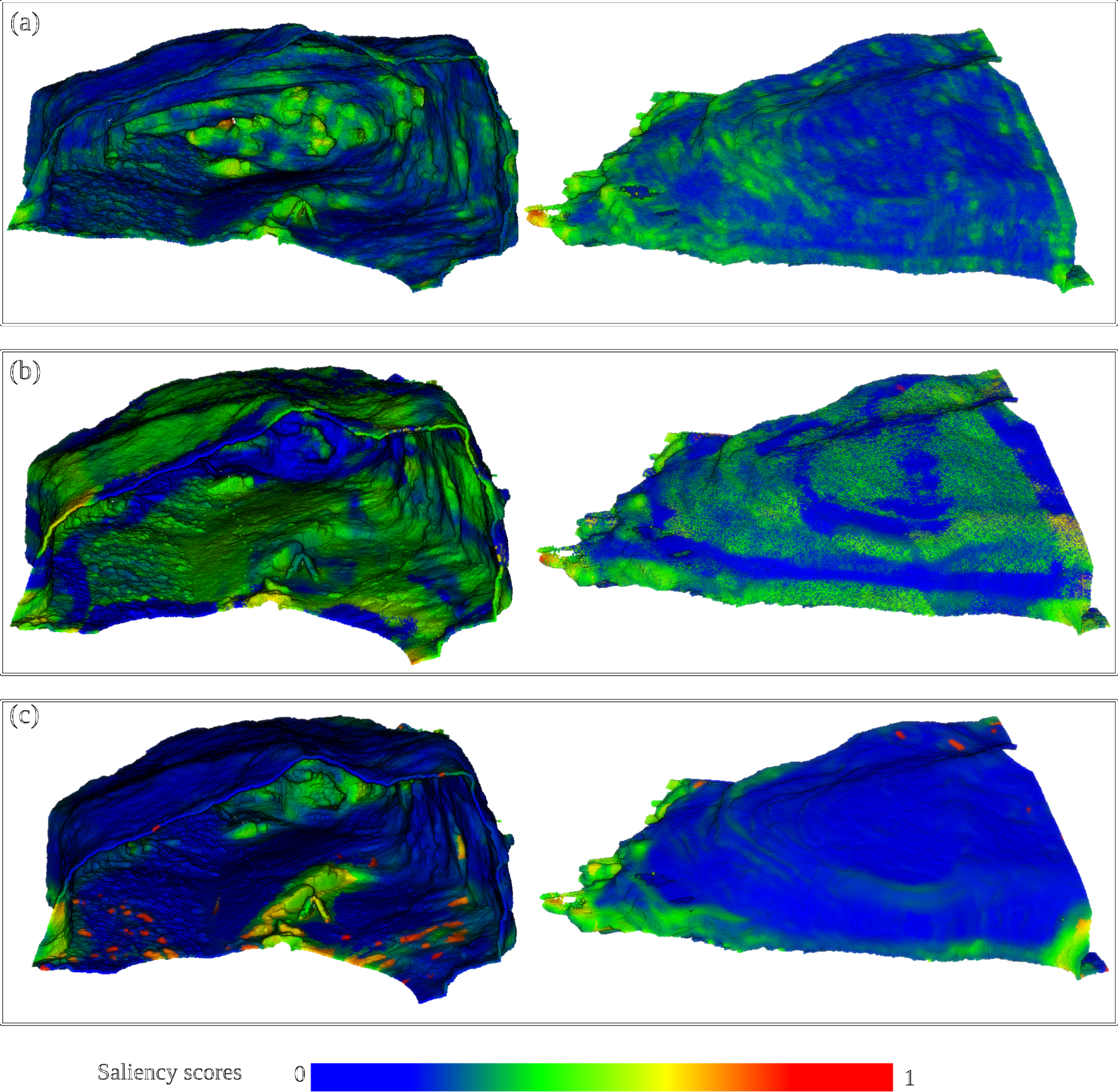}
	\caption{saliency scores using the three methods on the terrestrial dataset. Horizontal look into the cave (left) and at the walls and ceiling (right). Note that the ceiling point cloud was acquired from within the cave. 
		(a) proposed method using $n=16$ and $f=64$; 
		(b) plane-based reconstruction using $f=24$
		(c) handcrafted method using $\rho =0.3$~m.}
	\label{fig:cavecompare}
\end{figure}

\Frefsub{fig:cavecompare}{a} shows the saliency map using the proposed method both inside the cave (left) and on the ceiling (right). 
It can be seen that the points which received higher saliency scores mostly belong to niches and pockets in the walls. Additionally, points that lie on some larger rocks also have higher saliency scores, as well as a tripod that stands close to the entrance. 
Points belonging to  blocks on the floor near the entrance were estimated with lower saliency scores.  This is probably due to the fact that they cover a large part of the cave floor. Therefore, they are considered as roughness that can be predicted by the proposed model. 

We used $n=16$ for the plane-based method and a minimal object size of 0.1~m for the handcrafted approach. It can be seen that for the plane-based method,  regions that deviate from planarity, which compose the majority of the dataset, were given higher scores  (\Fref{fig:cavecompare}, b).
The handcrafted method \PFrefsub{fig:cavecompare}{c} provided less noisy saliency map. Most of the rocks on the ground have lower saliency scores, similar to the proposed method.  
However, much less points that belong to niches in the ceiling were estimated with high saliency scores compared to  the proposed method. This leads to much more focused areas of interest. 

Looking at the saliency ratio in the test data \PTref{tab:rt_comparison}, it can be seen that the handcrafted method achieved the highest values by far, whereas the plane-based method yielded the lowest.
This is in accordance with the visual impression.

\section{Conclusions}
In this paper we proposed an unsupervised method that highlights saliency in non-urban, natural environments. 
Driven by the notion that salient regions stand out in their environment and knowing that topography is generally smooth,  we search for anomalies within a scanned surface. 
The proposed approach is trained to reconstruct the surface based on voxel grids extracted from the data.  
Based on training, it reconstructs the local surface and evaluates the difference between the inferred surface and the original point cloud. 
Saliency scores are defined based on the difference from the expected surface. Therefore, the network should be trained for every dataset. However, the model requires some examples for salient and non-salient areas in order to tune the hyper-parameters. Nevertheless, these samples are not required for the learning process per se. 

The proposed method was demonstrated on three datasets acquired by various scanning platforms in different types of scenes and presented  three levels of surface complexity (from smooth, almost planar surface, to rough riverbed and to a complex 3D cave).  
We have shown that it  was able to discern between `salient' and `non-salient' regions, yielding high saliency ratio. 

For evaluation, we proposed a saliency ratio metric, which measures the ratio between regions previously known to have higher and lower salient scores. In addition, we visually inspected the results, while comparing them to other baseline approaches of saliency detection. 
We have shown that in most cases, the propose metric corresponds to the visual results. 

Further examination into the more important hyper-parameters, $f$ and $n$, revealed that the size of the voxel grid dictates the size of the detected region. As $ n $ increases, a larger region is reconstructed, and evidently, larger parts will deviate from the original cloud. This will result in generally higher saliency scores. Though the number of feature maps in base resolution is important to reconstruct the surface, its effect is limited. Nonetheless, a  sufficient number of feature maps in base resolution is required for the reconstruction. Too many, or too less maps, will lead to higher saliency scores, also in non-salient regions. 
That said, we have shown that the effect of these parameters on the final results (both visually and quantitatively) is limited, especially when the surface is more complex (the cave, as an example). 

When compared to baseline methods, the handcrafted approach showed some advantage over the proposed method, as it delivered  more focused results. However, we have shown that its results highly depend on the scanning pattern. When point density was changing drastically (dataset \#II, for example), the handcrafted method estimated high saliency scores in non-salient regions. In contrast, the proposed method was unaffected, and showed similar results independently.



\section*{Acknowledgments}
This work was supported in part by the European
Union’s Horizon 2020 Research and Innovation Programme under the Marie
Skłodowska-Curie Grant 896409. The authors would also like to thank Prof. Gottfried Mandlburger for sharing dataset \#II.

\bibliographystyle{elsarticle-harv}
\bibliography{00_Anomaly}

\begin{thebibliography}{53}
\expandafter\ifx\csname natexlab\endcsname\relax\def\natexlab#1{#1}\fi
\providecommand{\url}[1]{\texttt{#1}}
\providecommand{\href}[2]{#2}
\providecommand{\path}[1]{#1}
\providecommand{\DOIprefix}{doi:}
\providecommand{\ArXivprefix}{arXiv:}
\providecommand{\URLprefix}{URL: }
\providecommand{\Pubmedprefix}{pmid:}
\providecommand{\doi}[1]{\href{http://dx.doi.org/#1}{\path{#1}}}
\providecommand{\Pubmed}[1]{\href{pmid:#1}{\path{#1}}}
\providecommand{\bibinfo}[2]{#2}
\ifx\xfnm\relax \def\xfnm[#1]{\unskip,\space#1}\fi
\bibitem[{Akman and Jonker(2010)}]{Akman.Jonker2010}
\bibinfo{author}{Akman, O.}, \bibinfo{author}{Jonker, P.},
  \bibinfo{year}{2010}.
\newblock \bibinfo{title}{Computing saliency map from spatial information in
  point cloud data}, in: \bibinfo{booktitle}{Advanced Concepts for Intelligent
  Vision Systems}. \bibinfo{publisher}{Springer Berlin Heidelberg}, pp.
  \bibinfo{pages}{290--299}.
\newblock \DOIprefix\doi{10.1007/978-3-642-17688-3\_28}.
\bibitem[{Alexiou et~al.(2019)Alexiou, Xu and Ebrahimi}]{Alexiou.etal2019}
\bibinfo{author}{Alexiou, E.}, \bibinfo{author}{Xu, P.},
  \bibinfo{author}{Ebrahimi, T.}, \bibinfo{year}{2019}.
\newblock \bibinfo{title}{Towards modelling of visual saliency in point clouds
  for immersive applications}, in: \bibinfo{booktitle}{2019 {IEEE}
  International Conference on Image Processing ({ICIP})}, pp.
  \bibinfo{pages}{4325--4329}.
\newblock \DOIprefix\doi{10.1109/icip.2019.8803479}.
\bibitem[{Arav and Filin(2020)}]{Arav.Filin2020}
\bibinfo{author}{Arav, R.}, \bibinfo{author}{Filin, S.}, \bibinfo{year}{2020}.
\newblock \bibinfo{title}{Saliency of subtle entities within {3-D} point
  clouds}, \bibinfo{publisher}{Copernicus {GmbH}}. pp.
  \bibinfo{pages}{179--186}.
\newblock \DOIprefix\doi{10.5194/isprs-annals-v-2-2020-179-2020}.
\bibitem[{Arav and Filin(2022)}]{Arav.Filin2022}
\bibinfo{author}{Arav, R.}, \bibinfo{author}{Filin, S.}, \bibinfo{year}{2022}.
\newblock \bibinfo{title}{A visual saliency-driven extraction framework of
  smoothly embedded entities in {3D} point clouds of open terrain}.
\newblock \bibinfo{journal}{{ISPRS} Journal of Photogrammetry and Remote
  Sensing} \bibinfo{volume}{188}, \bibinfo{pages}{125--140}.
\newblock \DOIprefix\doi{10.1016/j.isprsjprs.2022.04.003}.
\bibitem[{Arav et~al.(2022a)Arav, Filin and Pfeifer}]{Arav.etal2022}
\bibinfo{author}{Arav, R.}, \bibinfo{author}{Filin, S.},
  \bibinfo{author}{Pfeifer, N.}, \bibinfo{year}{2022}a.
\newblock \bibinfo{title}{Content-aware point cloud simplification of open
  scenes}.
\newblock \bibinfo{journal}{IEEE Transactions on Geoscience and Remote Sensing}
  \bibinfo{volume}{60}, \bibinfo{pages}{1--12}.
\newblock \DOIprefix\doi{10.1109/TGRS.2022.3208348}.
\bibitem[{Arav et~al.(2022b)Arav, Pöppl and Pfeifer}]{Arav.etal2022a}
\bibinfo{author}{Arav, R.}, \bibinfo{author}{Pöppl, F.},
  \bibinfo{author}{Pfeifer, N.}, \bibinfo{year}{2022}b.
\newblock \bibinfo{title}{A point-based level-set approach for the extraction
  of {3D} entities from point clouds -- application in geomorphological
  context}, \bibinfo{publisher}{Copernicus {GmbH}}. pp.
  \bibinfo{pages}{95--102}.
\newblock \DOIprefix\doi{10.5194/isprs-annals-v-2-2022-95-2022}.
\bibitem[{Arav et~al.(2024)Arav, Ressl, Weiss, Artz and
  Mandlburger}]{Arav.etal2024}
\bibinfo{author}{Arav, R.}, \bibinfo{author}{Ressl, C.},
  \bibinfo{author}{Weiss, R.}, \bibinfo{author}{Artz, T.},
  \bibinfo{author}{Mandlburger, G.}, \bibinfo{year}{2024}.
\newblock \bibinfo{title}{Evaluation of active and passive uav-based surveying
  systems for eulittoral zone mapping}.
\newblock \bibinfo{journal}{The International Archives of the Photogrammetry,
  Remote Sensing and Spatial Information Sciences}
  \bibinfo{volume}{XLVIII-2–2024}, \bibinfo{pages}{9--16}.
\newblock \DOIprefix\doi{10.5194/isprs-archives-xlviii-2-2024-9-2024}.
\bibitem[{Arav and Wittich()}]{code2023}
\bibinfo{author}{Arav, R.}, \bibinfo{author}{Wittich, D.}, .
\newblock \bibinfo{title}{Salient\_anomaly}.
\newblock \URLprefix
  \url{https://github.com/rarav/salient_anomaly/releases/tag/v1.0.0}.
\bibitem[{Arvanitis et~al.(2022)Arvanitis, Zacharaki, Vasa and
  Moustakas}]{Arvanitis.etal2022}
\bibinfo{author}{Arvanitis, G.}, \bibinfo{author}{Zacharaki, E.I.},
  \bibinfo{author}{Vasa, L.}, \bibinfo{author}{Moustakas, K.},
  \bibinfo{year}{2022}.
\newblock \bibinfo{title}{Broad-to-narrow registration and identification of
  {3D} objects in partially scanned and cluttered point clouds}.
\newblock \bibinfo{journal}{{IEEE} Transactions on Multimedia}
  \bibinfo{volume}{24}, \bibinfo{pages}{2230--2245}.
\newblock \DOIprefix\doi{10.1109/tmm.2021.3089838}.
\bibitem[{Chen et~al.(2021a)Chen, Li, Deng and Lin}]{Chen.etal2021}
\bibinfo{author}{Chen, H.}, \bibinfo{author}{Li, Y.}, \bibinfo{author}{Deng,
  Y.}, \bibinfo{author}{Lin, G.}, \bibinfo{year}{2021}a.
\newblock \bibinfo{title}{{CNN}-based {RGB}-d salient object detection: Learn,
  select, and fuse}.
\newblock \bibinfo{journal}{International Journal of Computer Vision}
  \bibinfo{volume}{129}, \bibinfo{pages}{2076--2096}.
\newblock \DOIprefix\doi{10.1007/s11263-021-01452-0}.
\bibitem[{Chen et~al.(2021b)Chen, Fu, Liu, Chen, Du, Qiu and
  Shao}]{Chen.etal2021a}
\bibinfo{author}{Chen, Q.}, \bibinfo{author}{Fu, K.}, \bibinfo{author}{Liu,
  Z.}, \bibinfo{author}{Chen, G.}, \bibinfo{author}{Du, H.},
  \bibinfo{author}{Qiu, B.}, \bibinfo{author}{Shao, L.}, \bibinfo{year}{2021}b.
\newblock \bibinfo{title}{{EF}-net: A novel enhancement and fusion network for
  {RGB}-d saliency detection}.
\newblock \bibinfo{journal}{Pattern Recognition} \bibinfo{volume}{112},
  \bibinfo{pages}{107740}.
\newblock \DOIprefix\doi{10.1016/j.patcog.2020.107740}.
\bibitem[{Ding et~al.(2019)Ding, Lin, Chen and Zhang}]{Ding.etal2019}
\bibinfo{author}{Ding, X.}, \bibinfo{author}{Lin, W.}, \bibinfo{author}{Chen,
  Z.}, \bibinfo{author}{Zhang, X.}, \bibinfo{year}{2019}.
\newblock \bibinfo{title}{Point cloud saliency detection by local and global
  feature fusion}.
\newblock \bibinfo{journal}{{IEEE} Transactions on Image Processing} ,
  \bibinfo{pages}{1--1}\DOIprefix\doi{10.1109/tip.2019.2918735}.
\bibitem[{Fan et~al.(2022)Fan, Gao and Li}]{Fan.etal2022}
\bibinfo{author}{Fan, S.}, \bibinfo{author}{Gao, W.}, \bibinfo{author}{Li, G.},
  \bibinfo{year}{2022}.
\newblock \bibinfo{title}{Salient object detection for~point clouds}, in:
  \bibinfo{booktitle}{Lecture Notes in Computer Science}.
  \bibinfo{publisher}{Springer Nature Switzerland}, pp. \bibinfo{pages}{1--19}.
\newblock \DOIprefix\doi{10.1007/978-3-031-19815-1\_1}.
\bibitem[{{Geological Survey of Israel} and Arav(2013)}]{GSI.Arav2013}
[dataset] \bibinfo{author}{{Geological Survey of Israel}},
  \bibinfo{author}{Arav, R.}, \bibinfo{year}{2013}.
\newblock \bibinfo{title}{Zeelim fan, israel (part)}.
\newblock \bibinfo{howpublished}{TU Data Repository}.
\newblock \DOIprefix\doi{10.48436/mps0m-c9n43}. \bibinfo{note}{[online]}.
\bibitem[{Guo et~al.(2018)Guo, Wang and Xin}]{Guo.etal2018}
\bibinfo{author}{Guo, Y.}, \bibinfo{author}{Wang, F.}, \bibinfo{author}{Xin,
  J.}, \bibinfo{year}{2018}.
\newblock \bibinfo{title}{Point-wise saliency detection on {3D} point clouds
  via covariance descriptors}.
\newblock \bibinfo{journal}{Visual Computer} \bibinfo{volume}{34},
  \bibinfo{pages}{1325--1338}.
\newblock \DOIprefix\doi{10.1007/s00371-017-1416-3}.
\bibitem[{Han et~al.(2018)Han, Chen, Liu, Yan and Li}]{Han.etal2018}
\bibinfo{author}{Han, J.}, \bibinfo{author}{Chen, H.}, \bibinfo{author}{Liu,
  N.}, \bibinfo{author}{Yan, C.}, \bibinfo{author}{Li, X.},
  \bibinfo{year}{2018}.
\newblock \bibinfo{title}{{CNNs}-based {RGB}-d saliency detection via
  cross-view transfer and multiview fusion}.
\newblock \bibinfo{journal}{{IEEE} Transactions on Cybernetics}
  \bibinfo{volume}{48}, \bibinfo{pages}{3171--3183}.
\newblock \DOIprefix\doi{10.1109/tcyb.2017.2761775}.
\bibitem[{Hao et~al.(2019)Hao, Liang, Wang, Zhao and Li}]{Hao.etal2019}
\bibinfo{author}{Hao, W.}, \bibinfo{author}{Liang, W.}, \bibinfo{author}{Wang,
  Y.}, \bibinfo{author}{Zhao, M.}, \bibinfo{author}{Li, Y.},
  \bibinfo{year}{2019}.
\newblock \bibinfo{title}{Saliency-guided repetition detection from facade
  point clouds}.
\newblock \bibinfo{journal}{{IEEE} Access} \bibinfo{volume}{7},
  \bibinfo{pages}{150072--150081}.
\newblock \DOIprefix\doi{10.1109/access.2019.2947537}.
\bibitem[{Hillier et~al.(2014)Hillier, Smith, Armugam, Barr, Boston, Clark,
  Ely, Frankl, Greenwood, Gosselin, H{\"{a}}ttestrand, Hogan, Hughes,
  Livingstone, Lovell, McHenry, Munoz, Pellicer, Pellitero, Robb, Roberson,
  Ruther, Spagnolo, Standell, Stokes, Storrar, Tate and
  Wooldridge}]{Hillier.etal2014}
\bibinfo{author}{Hillier, J.K.}, \bibinfo{author}{Smith, M.J.},
  \bibinfo{author}{Armugam, R.}, \bibinfo{author}{Barr, I.},
  \bibinfo{author}{Boston, C.M.}, \bibinfo{author}{Clark, C.D.},
  \bibinfo{author}{Ely, J.}, \bibinfo{author}{Frankl, A.},
  \bibinfo{author}{Greenwood, S.L.}, \bibinfo{author}{Gosselin, L.},
  \bibinfo{author}{H{\"{a}}ttestrand, C.}, \bibinfo{author}{Hogan, K.},
  \bibinfo{author}{Hughes, A.L.C.}, \bibinfo{author}{Livingstone, S.J.},
  \bibinfo{author}{Lovell, H.}, \bibinfo{author}{McHenry, M.},
  \bibinfo{author}{Munoz, Y.}, \bibinfo{author}{Pellicer, X.M.},
  \bibinfo{author}{Pellitero, R.}, \bibinfo{author}{Robb, C.},
  \bibinfo{author}{Roberson, S.}, \bibinfo{author}{Ruther, D.},
  \bibinfo{author}{Spagnolo, M.}, \bibinfo{author}{Standell, M.},
  \bibinfo{author}{Stokes, C.R.}, \bibinfo{author}{Storrar, R.},
  \bibinfo{author}{Tate, N.J.}, \bibinfo{author}{Wooldridge, K.},
  \bibinfo{year}{2014}.
\newblock \bibinfo{title}{{Manual mapping of drumlins in synthetic landscapes
  to assess operator effectiveness}}.
\newblock \bibinfo{journal}{Journal of Maps} \bibinfo{volume}{11},
  \bibinfo{pages}{719--729}.
\newblock \URLprefix
  \url{https://www.tandfonline.com/doi/full/10.1080/17445647.2014.957251},
  \DOIprefix\doi{10.1080/17445647.2014.957251}.
\bibitem[{Hong et~al.(2023)Hong, Zhang and Ma}]{Hong.etal2023}
\bibinfo{author}{Hong, T.}, \bibinfo{author}{Zhang, Z.}, \bibinfo{author}{Ma,
  J.}, \bibinfo{year}{2023}.
\newblock \bibinfo{title}{{PCSalmix}: Gradient saliency-based mix augmentation
  for point cloud classification}, in: \bibinfo{booktitle}{{ICASSP} 2023 - 2023
  {IEEE} International Conference on Acoustics, Speech and Signal Processing
  ({ICASSP})}, \bibinfo{publisher}{{IEEE}}.
\newblock \DOIprefix\doi{10.1109/icassp49357.2023.10095576}.
\bibitem[{Itti et~al.(1998)Itti, Koch and Niebur}]{Itti.etal1998}
\bibinfo{author}{Itti, L.}, \bibinfo{author}{Koch, C.},
  \bibinfo{author}{Niebur, E.}, \bibinfo{year}{1998}.
\newblock \bibinfo{title}{A model of saliency-based visual attention for rapid
  scene analysis}.
\newblock \bibinfo{journal}{IEEE Transactions on Pattern Analysis and Machine
  Intelligence} \bibinfo{volume}{20}, \bibinfo{pages}{1254--1259}.
\newblock \DOIprefix\doi{10.1109/34.730558}.
\bibitem[{Jiang et~al.(2023)Jiang, Ding, Tam, Song, Li and
  Yang}]{Jiang.etal2023}
\bibinfo{author}{Jiang, Z.}, \bibinfo{author}{Ding, L.}, \bibinfo{author}{Tam,
  G.K.}, \bibinfo{author}{Song, C.}, \bibinfo{author}{Li, F.W.},
  \bibinfo{author}{Yang, B.}, \bibinfo{year}{2023}.
\newblock \bibinfo{title}{C2spoint: A classification-to-saliency network for
  point cloud saliency detection}.
\newblock \bibinfo{journal}{Computers \& Graphics} \bibinfo{volume}{115},
  \bibinfo{pages}{274--284}.
\newblock \DOIprefix\doi{10.1016/j.cag.2023.07.003}.
\bibitem[{Kobyshev et~al.(2016)Kobyshev, Riemenschneider,
  B{\'{o}}dis-Szomor{\'{u}} and Gool}]{Kobyshev.etal2016}
\bibinfo{author}{Kobyshev, N.}, \bibinfo{author}{Riemenschneider, H.},
  \bibinfo{author}{B{\'{o}}dis-Szomor{\'{u}}, A.}, \bibinfo{author}{Gool,
  L.V.}, \bibinfo{year}{2016}.
\newblock \bibinfo{title}{{3D} saliency for finding landmark buildings}, in:
  \bibinfo{booktitle}{2016 Fourth International Conference on 3D Vision (3DV)},
  pp. \bibinfo{pages}{267--275}.
\newblock \DOIprefix\doi{10.1109/3DV.2016.35}.
\bibitem[{Kyriou et~al.(2021)Kyriou, Nikolakopoulos and
  Koukouvelas}]{Kyriou.etal2021}
\bibinfo{author}{Kyriou, A.}, \bibinfo{author}{Nikolakopoulos, K.},
  \bibinfo{author}{Koukouvelas, I.}, \bibinfo{year}{2021}.
\newblock \bibinfo{title}{How image acquisition geometry of {UAV} campaigns
  affects the derived products and their accuracy in areas with complex
  geomorphology}.
\newblock \bibinfo{journal}{{ISPRS} International Journal of Geo-Information}
  \bibinfo{volume}{10}, \bibinfo{pages}{408}.
\newblock \DOIprefix\doi{10.3390/ijgi10060408}.
\bibitem[{Laazoufi and Hassouni(2022)}]{Laazoufi.Hassouni2022}
\bibinfo{author}{Laazoufi, A.}, \bibinfo{author}{Hassouni, M.E.},
  \bibinfo{year}{2022}.
\newblock \bibinfo{title}{Saliency-based point cloud quality assessment method
  using aware features learning}, in: \bibinfo{booktitle}{2022 9th
  International Conference on Wireless Networks and Mobile Communications
  ({WINCOM})}, \bibinfo{publisher}{{IEEE}}.
\newblock \DOIprefix\doi{10.1109/wincom55661.2022.9966464}.
\bibitem[{Leal et~al.(2019)Leal, Sanchez-Torres and
  Branch-Bedoya}]{Leal.etal2019}
\bibinfo{author}{Leal, E.A.}, \bibinfo{author}{Sanchez-Torres, G.},
  \bibinfo{author}{Branch-Bedoya, J.W.}, \bibinfo{year}{2019}.
\newblock \bibinfo{title}{Point cloud saliency detection via local sparse
  coding}.
\newblock \bibinfo{journal}{{DYNA}} \bibinfo{volume}{86},
  \bibinfo{pages}{238--247}.
\newblock \DOIprefix\doi{10.15446/dyna.v86n209.75958}.
\bibitem[{Li et~al.(2022)Li, Ji, Zhang, Piao, Lu and Cheng}]{Li.etal2022}
\bibinfo{author}{Li, J.}, \bibinfo{author}{Ji, W.}, \bibinfo{author}{Zhang,
  M.}, \bibinfo{author}{Piao, Y.}, \bibinfo{author}{Lu, H.},
  \bibinfo{author}{Cheng, L.}, \bibinfo{year}{2022}.
\newblock \bibinfo{title}{Delving into calibrated depth for accurate {RGB}-d
  salient object detection}.
\newblock \bibinfo{journal}{International Journal of Computer Vision}
  \bibinfo{volume}{131}, \bibinfo{pages}{855--876}.
\newblock \DOIprefix\doi{10.1007/s11263-022-01734-1}.
\bibitem[{Liang et~al.(2023)Liang, Zhang, Hua and Chen}]{Liang.etal2023}
\bibinfo{author}{Liang, A.}, \bibinfo{author}{Zhang, H.}, \bibinfo{author}{Hua,
  H.}, \bibinfo{author}{Chen, W.}, \bibinfo{year}{2023}.
\newblock \bibinfo{title}{To drop or to select: Reduce the negative effects of
  disturbance features for point cloud classification from an interpretable
  perspective}.
\newblock \bibinfo{journal}{{IEEE} Access} \bibinfo{volume}{11},
  \bibinfo{pages}{36184--36202}.
\newblock \DOIprefix\doi{10.1109/access.2023.3266340}.
\bibitem[{Liu et~al.(2019)Liu, Yuan and Gao}]{Liu.etal2019}
\bibinfo{author}{Liu, Y.}, \bibinfo{author}{Yuan, Y.}, \bibinfo{author}{Gao,
  S.}, \bibinfo{year}{2019}.
\newblock \bibinfo{title}{Modeling the vagueness of areal geographic objects: A
  categorization system}.
\newblock \bibinfo{journal}{ISPRS International Journal of Geo-Information}
  \bibinfo{volume}{8}, \bibinfo{pages}{306}.
\newblock \DOIprefix\doi{10.3390/ijgi8070306}.
\bibitem[{Mandlburger et~al.(2015)Mandlburger, Hauer, Wieser and
  Pfeifer}]{Mandlburger.etal2015}
\bibinfo{author}{Mandlburger, G.}, \bibinfo{author}{Hauer, C.},
  \bibinfo{author}{Wieser, M.}, \bibinfo{author}{Pfeifer, N.},
  \bibinfo{year}{2015}.
\newblock \bibinfo{title}{Topo-bathymetric {LiDAR} for monitoring river
  morphodynamics and instream habitats{\textemdash}a case study at the pielach
  river}.
\newblock \bibinfo{journal}{Remote Sensing} \bibinfo{volume}{7},
  \bibinfo{pages}{6160--6195}.
\newblock \DOIprefix\doi{10.3390/rs70506160}.
\bibitem[{Molenaar and Cheng(2000)}]{Molenaar.Cheng2000}
\bibinfo{author}{Molenaar, M.}, \bibinfo{author}{Cheng, T.},
  \bibinfo{year}{2000}.
\newblock \bibinfo{title}{Fuzzy spatial objects and their dynamics}.
\newblock \bibinfo{journal}{{ISPRS} Journal of Photogrammetry and Remote
  Sensing} \bibinfo{volume}{55}, \bibinfo{pages}{164--175}.
\newblock \DOIprefix\doi{10.1016/s0924-2716(00)00017-4}.
\bibitem[{Pfeifer and Mandlburger(2018)}]{Pfeifer.Mandlburger2018}
\bibinfo{author}{Pfeifer, N.}, \bibinfo{author}{Mandlburger, G.},
  \bibinfo{year}{2018}.
\newblock \bibinfo{title}{Topographic Laser Ranging and Scanning}.
  \bibinfo{publisher}{{CRC} Press}. chapter \bibinfo{chapter}{LiDAR Data
  Filtering and Digital Terrain Model Generation}.
\newblock p.~\bibinfo{pages}{30}.
\newblock \DOIprefix\doi{10.1201/9781315154381}.
\bibitem[{Pfeifer et~al.(2014)Pfeifer, Mandlburger, Otepka and
  Karel}]{Pfeifer.etal2014}
\bibinfo{author}{Pfeifer, N.}, \bibinfo{author}{Mandlburger, G.},
  \bibinfo{author}{Otepka, J.}, \bibinfo{author}{Karel, W.},
  \bibinfo{year}{2014}.
\newblock \bibinfo{title}{{OPALS} {\textendash} a framework for airborne laser
  scanning data analysis}.
\newblock \bibinfo{journal}{Computers, Environment and Urban Systems}
  \bibinfo{volume}{45}, \bibinfo{pages}{125--136}.
\newblock \DOIprefix\doi{10.1016/j.compenvurbsys.2013.11.002}.
\bibitem[{Qin et~al.(2023)Qin, Wang, Peng and Xu}]{Qin.etal2023}
\bibinfo{author}{Qin, Z.}, \bibinfo{author}{Wang, C.}, \bibinfo{author}{Peng,
  Y.}, \bibinfo{author}{Xu, K.}, \bibinfo{year}{2023}.
\newblock \bibinfo{title}{{CasViGE}: Learning robust point cloud registration
  with cascaded visual-geometric encoding}.
\newblock \bibinfo{journal}{Computer Aided Geometric Design}
  \bibinfo{volume}{104}, \bibinfo{pages}{102217}.
\newblock \DOIprefix\doi{10.1016/j.cagd.2023.102217}.
\bibitem[{Ronneberger et~al.(2015)Ronneberger, Fischer and
  Brox}]{ronneberger2015}
\bibinfo{author}{Ronneberger, O.}, \bibinfo{author}{Fischer, P.},
  \bibinfo{author}{Brox, T.}, \bibinfo{year}{2015}.
\newblock \bibinfo{title}{U-net: Convolutional networks for biomedical image
  segmentation}, in: \bibinfo{booktitle}{Medical Image Computing and
  Computer-Assisted Intervention--MICCAI 2015: 18th International Conference,
  Munich, Germany, October 5-9, 2015, Proceedings, Part III 18},
  \bibinfo{organization}{Springer}. pp. \bibinfo{pages}{234--241}.
\bibitem[{Rusu et~al.(2009)Rusu, Blodow and Beetz}]{Rusu.etal2009}
\bibinfo{author}{Rusu, R.B.}, \bibinfo{author}{Blodow, N.},
  \bibinfo{author}{Beetz, M.}, \bibinfo{year}{2009}.
\newblock \bibinfo{title}{Fast point feature histograms ({FPFH}) for {3D}
  registration}, in: \bibinfo{booktitle}{2009 {IEEE} International Conference
  on Robotics and Automation}, \bibinfo{publisher}{{IEEE}}. pp.
  \bibinfo{pages}{3212--3217}.
\newblock \DOIprefix\doi{10.1109/robot.2009.5152473}.
\bibitem[{Scheiber et~al.(2015)Scheiber, Fredin, Viola, Jarna, Gasser and
  {\L}api{\'{n}}ska-Viola}]{Scheiber.etal2015}
\bibinfo{author}{Scheiber, T.}, \bibinfo{author}{Fredin, O.},
  \bibinfo{author}{Viola, G.}, \bibinfo{author}{Jarna, A.},
  \bibinfo{author}{Gasser, D.}, \bibinfo{author}{{\L}api{\'{n}}ska-Viola, R.},
  \bibinfo{year}{2015}.
\newblock \bibinfo{title}{Manual extraction of bedrock lineaments from
  high-resolution {LiDAR} data: methodological bias and human perception}.
\newblock \bibinfo{journal}{{GFF}} \bibinfo{volume}{137},
  \bibinfo{pages}{362--372}.
\newblock \URLprefix
  \url{http://www.tandfonline.com/doi/full/10.1080/11035897.2015.1085434},
  \DOIprefix\doi{10.1080/11035897.2015.1085434}.
\bibitem[{Shtrom et~al.(2013)Shtrom, Leifman and Tal}]{Shtrom.etal2013}
\bibinfo{author}{Shtrom, E.}, \bibinfo{author}{Leifman, G.},
  \bibinfo{author}{Tal, A.}, \bibinfo{year}{2013}.
\newblock \bibinfo{title}{{Saliency Detection in Large Point Sets}}, in:
  \bibinfo{booktitle}{Computer Vision (ICCV), 2013 IEEE International
  Conference on}, pp. \bibinfo{pages}{3591--3598}.
\newblock \DOIprefix\doi{10.1109/ICCV.2013.446}.
\bibitem[{Sudre et~al.(2017)Sudre, Li, Vercauteren, Ourselin and
  Jorge~Cardoso}]{Sudre.etal2017}
\bibinfo{author}{Sudre, C.H.}, \bibinfo{author}{Li, W.},
  \bibinfo{author}{Vercauteren, T.}, \bibinfo{author}{Ourselin, S.},
  \bibinfo{author}{Jorge~Cardoso, M.}, \bibinfo{year}{2017}.
\newblock \bibinfo{title}{Generalised Dice Overlap as a Deep Learning Loss
  Function for Highly Unbalanced Segmentations}. \bibinfo{publisher}{Springer
  International Publishing}.
\newblock pp. \bibinfo{pages}{240--248}.
\newblock \DOIprefix\doi{10.1007/978-3-319-67558-9\_28}.
\bibitem[{Sun et~al.(2022)Sun, Zhang, Li, Guo, Song and Li}]{Sun.etal2022}
\bibinfo{author}{Sun, P.}, \bibinfo{author}{Zhang, W.}, \bibinfo{author}{Li,
  S.}, \bibinfo{author}{Guo, Y.}, \bibinfo{author}{Song, C.},
  \bibinfo{author}{Li, X.}, \bibinfo{year}{2022}.
\newblock \bibinfo{title}{Learnable depth-sensitive attention for deep {RGB}-d
  saliency detection with multi-modal fusion architecture search}.
\newblock \bibinfo{journal}{International Journal of Computer Vision}
  \bibinfo{volume}{130}, \bibinfo{pages}{2822--2841}.
\newblock \DOIprefix\doi{10.1007/s11263-022-01646-0}.
\bibitem[{Tarolli and Mudd(2020)}]{Tarolli.Mudd2020}
\bibinfo{author}{Tarolli, P.}, \bibinfo{author}{Mudd, S.M.},
  \bibinfo{year}{2020}.
\newblock \bibinfo{title}{Introduction to remote sensing of geomorphology}, in:
  \bibinfo{booktitle}{Developments in Earth Surface Processes}.
  \bibinfo{publisher}{Elsevier}, pp. \bibinfo{pages}{xiii--xv}.
\newblock \DOIprefix\doi{10.1016/b978-0-444-64177-9.09992-6}.
\bibitem[{Tasse et~al.(2015)Tasse, Kosinka and Dodgson}]{Tasse.etal2015}
\bibinfo{author}{Tasse, P.F.}, \bibinfo{author}{Kosinka, J.},
  \bibinfo{author}{Dodgson, N.}, \bibinfo{year}{2015}.
\newblock \bibinfo{title}{{Cluster-based point set saliency}}, in:
  \bibinfo{booktitle}{Proceedings of the IEEE international conference on
  computer vision}, pp. \bibinfo{pages}{163--171}.
\bibitem[{Telling et~al.(2017)Telling, Lyda, Hartzell and
  Glennie}]{Telling.etal2017}
\bibinfo{author}{Telling, J.}, \bibinfo{author}{Lyda, A.},
  \bibinfo{author}{Hartzell, P.}, \bibinfo{author}{Glennie, C.},
  \bibinfo{year}{2017}.
\newblock \bibinfo{title}{Review of earth science research using terrestrial
  laser scanning}.
\newblock \bibinfo{journal}{Earth Sci. Rev.} \bibinfo{volume}{169},
  \bibinfo{pages}{35--68}.
\newblock \DOIprefix\doi{10.1016/j.earscirev.2017.04.007}.
\bibitem[{Tinchev et~al.(2021)Tinchev, Penate-Sanchez and
  Fallon}]{Tinchev.etal2021}
\bibinfo{author}{Tinchev, G.}, \bibinfo{author}{Penate-Sanchez, A.},
  \bibinfo{author}{Fallon, M.}, \bibinfo{year}{2021}.
\newblock \bibinfo{title}{{SKD}: Keypoint detection for point clouds using
  saliency estimation}.
\newblock \bibinfo{journal}{{IEEE} Robotics and Automation Letters}
  \bibinfo{volume}{6}, \bibinfo{pages}{3785--3792}.
\newblock \DOIprefix\doi{10.1109/lra.2021.3065224}.
\bibitem[{Vinci et~al.(2016)Vinci, Todisco and Mannocchi}]{Vinci.etal2016}
\bibinfo{author}{Vinci, A.}, \bibinfo{author}{Todisco, F.},
  \bibinfo{author}{Mannocchi, F.}, \bibinfo{year}{2016}.
\newblock \bibinfo{title}{{Calibration of manual measurements of rills using
  Terrestrial Laser Scanning}}.
\newblock \bibinfo{journal}{CATENA} \bibinfo{volume}{140},
  \bibinfo{pages}{164--168}.
\newblock \URLprefix
  \url{https://www.sciencedirect.com/science/article/pii/S034181621630042X},
  \DOIprefix\doi{10.1016/J.CATENA.2016.01.026}.
\bibitem[{Wang et~al.(2015)Wang, Luo, Wen, Cheng, Li, Chen, Wang and
  Li}]{Wang.etal2015}
\bibinfo{author}{Wang, H.}, \bibinfo{author}{Luo, H.}, \bibinfo{author}{Wen,
  C.}, \bibinfo{author}{Cheng, J.}, \bibinfo{author}{Li, P.},
  \bibinfo{author}{Chen, Y.}, \bibinfo{author}{Wang, C.}, \bibinfo{author}{Li,
  J.}, \bibinfo{year}{2015}.
\newblock \bibinfo{title}{{Road boundaries detection based on local normal
  saliency from mobile laser scanning data}}.
\newblock \bibinfo{journal}{IEEE Geoscience and remote sensing letters}
  \bibinfo{volume}{12}, \bibinfo{pages}{2085--2089}.
\newblock \DOIprefix\doi{10.1109/LGRS.2015.2449074}.
\bibitem[{Wimmer and Oberender(2022)}]{Wimmer.Oberender2022}
[dataset] \bibinfo{author}{Wimmer, M.}, \bibinfo{author}{Oberender, P.},
  \bibinfo{year}{2022}.
\newblock \bibinfo{title}{Untere traisenbacher höhle - utb\_104423}.
\newblock \bibinfo{howpublished}{TU Data Repository}.
\newblock \DOIprefix\doi{10.48436/fh0am-at738}. \bibinfo{note}{[online]}.
\bibitem[{Xu et~al.(2020)Xu, Li, Du, Zhang and Liu}]{xu2020reluplex}
\bibinfo{author}{Xu, J.}, \bibinfo{author}{Li, Z.}, \bibinfo{author}{Du, B.},
  \bibinfo{author}{Zhang, M.}, \bibinfo{author}{Liu, J.}, \bibinfo{year}{2020}.
\newblock \bibinfo{title}{Reluplex made more practical: Leaky relu}, in:
  \bibinfo{booktitle}{2020 IEEE Symposium on Computers and communications
  (ISCC)}, \bibinfo{organization}{IEEE}. pp. \bibinfo{pages}{1--7}.
\bibitem[{Yun and Sim(2016)}]{Yun.Sim2016}
\bibinfo{author}{Yun, J.S.}, \bibinfo{author}{Sim, J.Y.}, \bibinfo{year}{2016}.
\newblock \bibinfo{title}{Supervoxel-based saliency detection for large-scale
  colored {3D} point cloud}, in: \bibinfo{booktitle}{2020 Asia-Pacific Signal
  and Information Processing Association Annual Summit and Conference (APSIPA
  ASC)}, \bibinfo{publisher}{IEEE}. pp. \bibinfo{pages}{193--1200}.
\bibitem[{Zhang et~al.(2021a)Zhang, Fan, Dai, Anwar, Saleh, Aliakbarian and
  Barnes}]{Zhang.etal2021}
\bibinfo{author}{Zhang, J.}, \bibinfo{author}{Fan, D.P.}, \bibinfo{author}{Dai,
  Y.}, \bibinfo{author}{Anwar, S.}, \bibinfo{author}{Saleh, F.},
  \bibinfo{author}{Aliakbarian, S.}, \bibinfo{author}{Barnes, N.},
  \bibinfo{year}{2021}a.
\newblock \bibinfo{title}{Uncertainty inspired {RGB}-d saliency detection}.
\newblock \bibinfo{journal}{{IEEE} Transactions on Pattern Analysis and Machine
  Intelligence} ,
  \bibinfo{pages}{1--1}\DOIprefix\doi{10.1109/tpami.2021.3073564}.
\bibitem[{Zhang et~al.(2020)Zhang, Fan, Dai, Anwar, Saleh, Zhang and
  Barnes}]{Zhang.etal2020a}
\bibinfo{author}{Zhang, J.}, \bibinfo{author}{Fan, D.P.}, \bibinfo{author}{Dai,
  Y.}, \bibinfo{author}{Anwar, S.}, \bibinfo{author}{Saleh, F.S.},
  \bibinfo{author}{Zhang, T.}, \bibinfo{author}{Barnes, N.},
  \bibinfo{year}{2020}.
\newblock \bibinfo{title}{Uc-net: Uncertainty inspired rgb-d saliency detection
  via conditional variational autoencoders}, in:
  \bibinfo{booktitle}{Proceedings of the IEEE/CVF Conference on Computer Vision
  and Pattern Recognition (CVPR)}.
\bibitem[{Zhang et~al.(2021b)Zhang, Fan, Dai, Yu, Zhong, Barnes and
  Shao}]{Zhang.etal2021a}
\bibinfo{author}{Zhang, J.}, \bibinfo{author}{Fan, D.P.}, \bibinfo{author}{Dai,
  Y.}, \bibinfo{author}{Yu, X.}, \bibinfo{author}{Zhong, Y.},
  \bibinfo{author}{Barnes, N.}, \bibinfo{author}{Shao, L.},
  \bibinfo{year}{2021}b.
\newblock \bibinfo{title}{Rgb-d saliency detection via cascaded mutual
  information minimization}, in: \bibinfo{booktitle}{Proceedings of the
  IEEE/CVF International Conference on Computer Vision (ICCV)}, pp.
  \bibinfo{pages}{4338--4347}.
\bibitem[{Zheng et~al.(2019)Zheng, Chen, Yuan, Li and Ren}]{Zheng.etal2019}
\bibinfo{author}{Zheng, T.}, \bibinfo{author}{Chen, C.}, \bibinfo{author}{Yuan,
  J.}, \bibinfo{author}{Li, B.}, \bibinfo{author}{Ren, K.},
  \bibinfo{year}{2019}.
\newblock \bibinfo{title}{Pointcloud saliency maps}, in:
  \bibinfo{booktitle}{Proceedings of the IEEE/CVF International Conference on
  Computer Vision (ICCV)}, pp. \bibinfo{pages}{1598--1606}.
\bibitem[{Zhou et~al.(2021)Zhou, Lv, Lei and Yu}]{Zhou.etal2021}
\bibinfo{author}{Zhou, W.}, \bibinfo{author}{Lv, Y.}, \bibinfo{author}{Lei,
  J.}, \bibinfo{author}{Yu, L.}, \bibinfo{year}{2021}.
\newblock \bibinfo{title}{Global and local-contrast guides content-aware fusion
  for {RGB}-d saliency prediction}.
\newblock \bibinfo{journal}{{IEEE} Transactions on Systems, Man, and
  Cybernetics: Systems} \bibinfo{volume}{51}, \bibinfo{pages}{3641--3649}.
\newblock \DOIprefix\doi{10.1109/tsmc.2019.2957386}.

\end{thebibliography}

\end{document}